\documentclass[10pt,journal,compsoc]{IEEEtran}
\usepackage{graphicx}
\usepackage{epsfig}
\usepackage{graphicx}
\usepackage{amsmath}
\usepackage{amssymb}
\usepackage{graphicx}
\usepackage{comment}
\usepackage{amsmath,amssymb} 
\usepackage{color}
\usepackage{multirow}
\usepackage{epstopdf}
\usepackage{algorithm}
\usepackage{algorithmic}
\usepackage{booktabs}
\def \etal{{\emph{et al}}}
\usepackage{booktabs}
\usepackage{multirow}
\usepackage{algorithm}
\usepackage[colorlinks=true,citecolor=blue]{hyperref}

\usepackage{amsmath}
\usepackage{amssymb}
\usepackage{epsfig}
\usepackage{graphicx}
\usepackage{booktabs}
\usepackage{bbding}
\usepackage{pifont}

\usepackage{ragged2e}

\ifCLASSOPTIONcompsoc
  \usepackage[nocompress]{cite}
\else
  \usepackage{cite}
\fi
\ifCLASSINFOpdf
\else
\fi
\begin{document}
%
\title{Structural Prior Guided Generative Adversarial Transformers for Low-Light Image Enhancement }
%
%
%
%

\author{Cong Wang,
Jinshan Pan,
Xiao-Ming Wu
\IEEEcompsocitemizethanks{\IEEEcompsocthanksitem  Cong Wang, and Xiao-Ming Wu are with the Department of Computing, The Hong Kong Polytechnic University, Hong Kong, China (E-mail: supercong94@gmail.com, xiao-ming.wu@polyu.edu.hk).

\IEEEcompsocthanksitem Jinshan Pan is with the School of Computer Science and Engineering, Nanjing University of Science and Technology, Nanjing, China
(E-mail: sdluran@gmail.com)}

}

%
%

\markboth{Journal of \LaTeX\ Class Files,~Vol.~14, No.~8, August~2015}%
{Shell \MakeLowercase{\textit{et al.}}: Bare Demo of IEEEtran.cls for Computer Society Journals}
%



\IEEEtitleabstractindextext{%
\begin{abstract}
\justifying We propose an effective Structural Prior guided Generative Adversarial Transformer (SPGAT) to solve low-light image enhancement.
Our SPGAT mainly contains a generator with two discriminators and a structural prior estimator (SPE).
The generator is based on a U-shaped Transformer which is used to explore non-local information for better clear image restoration.
The SPE is used to explore useful structures from images to guide the generator for better structural detail estimation.
To generate more realistic images, we develop a new structural prior guided adversarial learning method by building the skip connections between the generator and discriminators so
that the discriminators can better discriminate between real and fake features.
Finally, we propose a parallel windows-based Swin Transformer block to aggregate different level hierarchical features for high-quality image restoration.
Experimental results demonstrate that the proposed SPGAT performs favorably against recent state-of-the-art methods on both synthetic and real-world datasets.
\end{abstract}

\begin{IEEEkeywords}
Low-light image enhancement, Transformer, Skip connections between generator and discriminator, Structural prior, Adversarial learning.
\end{IEEEkeywords}}

\maketitle

\IEEEdisplaynontitleabstractindextext

%
\IEEEpeerreviewmaketitle

\section{Introduction}\label{sec:introduction}

\IEEEPARstart{T}{aking} high-quality images in low-illumination environments is challenging as insufficient light usually leads to poor visibility that will affect further vision analysis and processing.
Thus, restoring a high-quality image from a given low-light image becomes a significantly important task.

Low-light Image Enhancement (LIE) is a challenging task as most of the important information in the images is missing.
To solve this problem, early approaches usually utilize histogram equalization~\cite{histogram-equalization_tce,dsp_histogram,yun2011histogram}, gamma correction~\cite{Gamma_tip,dsp_gamma}, and so on.
However, simply adjusting the pixel values does not effectively restore clear images.
Several methods~\cite{Retinex_tip15_fu,Retinex_tip18} formulate this problem by a Retinex model and develop kinds of effective image priors to solve this problem.
Although these approaches perform better than the histogram equalization-based ones, the designed priors are based on some statistical observations, which do not model the inherent properties of clear images well.

Deep learning, especially the deep convolutional neural network (CNN), provides an effective way to solve this problem.
Instead of designing sophisticated priors, these approaches usually directly estimate clear images from the low-light images via deep end-to-end trainable networks~\cite{mm19,mm20_lowlight,mm20_lv,eccv20,tcsvt_lowlight,tcsvt_RetinexDIP,tip_yang_drbn,tip_yang_retinex,guo_kindplus_ijcv,aaai22_jiang}.
As stated in~\cite{survey_li_pami}, the deep learning-based methods achieve better accuracy, robustness, and speed than conventional methods.

Although significant processes have been made, most existing deep learning methods do not restore structural details well as most of them do not model the structures of images in the network design.
As images usually contain rich structures which are vital for clear image restoration, it is of great interest to explore these structures to facilitate better structural detail restoration.
In addition, we note that most existing deep CNN-based methods mainly depend on local invariant convolution operations to extract features for image restoration, which does not model the non-local information.
As non-local image regions contain useful information, it is also of great interest to explore non-local information for low-light image enhancement.

\begin{figure*}[!t]
\begin{center}
\begin{tabular}{ccccc}
\includegraphics[width = 0.24\linewidth]{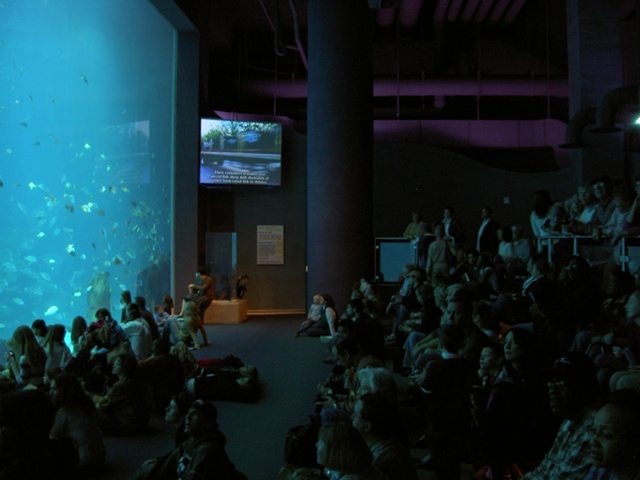}&\hspace{-4mm}
\includegraphics[width = 0.24\linewidth]{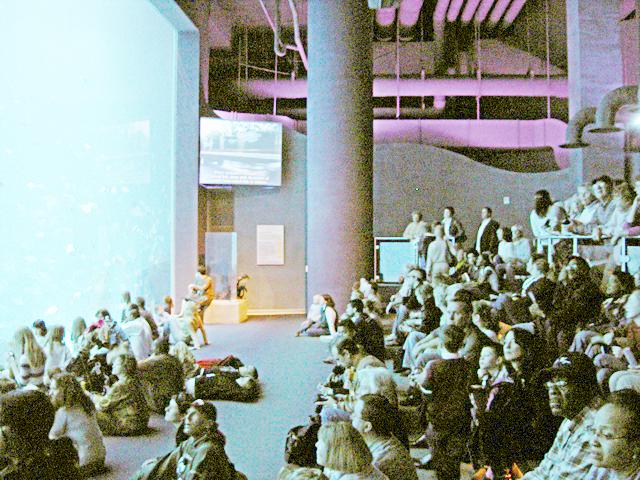}&\hspace{-4mm}
\includegraphics[width = 0.24\linewidth]{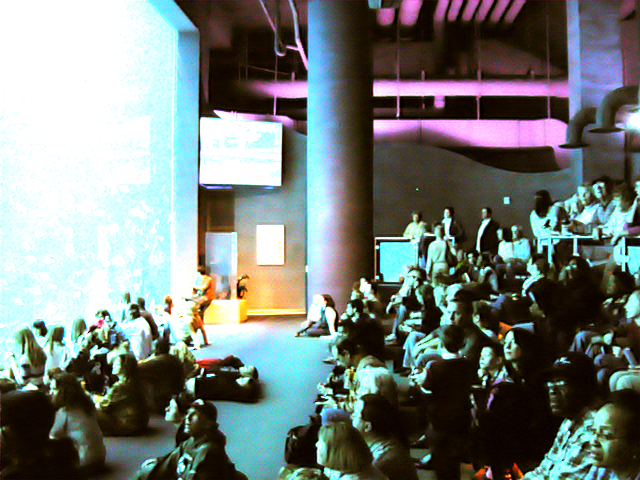}&\hspace{-4mm}
\includegraphics[width = 0.24\linewidth]{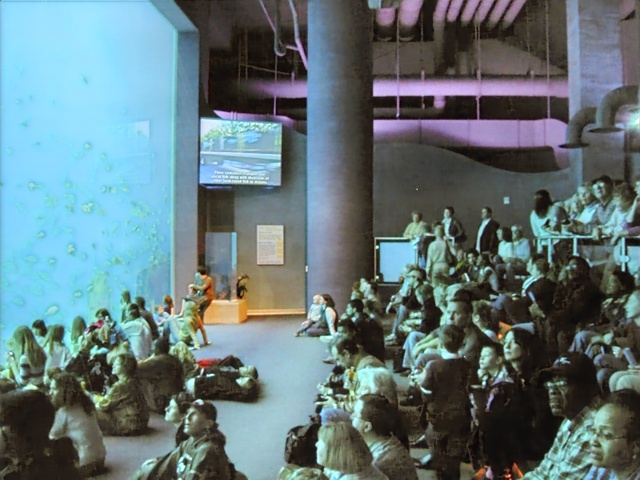}
\\
(a) Low-light &\hspace{-4mm} (b) Zero-DCE~\cite{zerodce_lowlight_guo} &\hspace{-4mm} (c) RUAS~\cite{RUAS_liu_cvpr21} &\hspace{-4mm} (d) SPGAT

\end{tabular}
\end{center}
\caption{Real-world low-light image enhancement results.
We develop an effective Structural Prior Guided Generative Adversarial Transformers (SPGAT) to better utilize the image structures and non-local information based on Transformers in a GAN framework for low-light image enhancement. As our analysis shows that our SPGAT can generate high-quality images with finer structural details than most existing methods.
}
\label{fig: introduction.}
\end{figure*}

To explore non-local information and useful structures of images for low-light image enhancement, we develop a Structural Prior guided Generative Adversarial Transformer (SPGAT) for low-light image enhancement.
First, we develop a generator based on a U-shaped Transformer with skip connections to explore non-local information for clear image restoration.
To restore high-quality images with structural details, we then propose a Structural Prior Estimator (SPE) to estimate structural features of images based on a U-shaped Transformer and develop an effective Structural Prior Guided Module (SPGM) to ensure that the estimated structural prior by SPE can better guide the generator for structural detail restoration.
Then, to generate more realistic images, we further develop a new structural prior guided adversarial learning method.
Specifically, we build the skip connections between the generator and discriminators so that the discriminators can better discriminate between real and fake features in the generator for more realistic features generation.
The image structure by the SPE is also utilized to guide the discriminators for better estimations.
Finally, we propose a parallel windows-based Swin Transformer block, which is the basic layer in generator, discriminators, and SPE, to aggregate different level hierarchical features for better enhancing images.
Fig.~\ref{sec:introduction} presents a real-world enhancement example compared with Zero-DCE~\cite{zerodce_lowlight_guo} and RUAS~\cite{RUAS_liu_cvpr21}, which shows that our method is able to generate a more natural result with better structural details.

The main contributions of our work are summarized as follows:
\begin{enumerate}
\item We propose a generator based on a U-shaped Transformer with skip connections to explore non-local information for clear image restoration.
\item We develop a simple and effective structural prior estimator to extract structural features from images to guide the estimations of the generator for structural detail estimation.
\item We propose a new structural prior guided adversarial learning manner by building the skip connections between the generator and discriminators so that the image structures from the generator can better constrain the discriminators for realistic image restoration.
\item We propose a parallel windows-based Swin Transformer block to better improve the quality of the restored images. Experiments demonstrate that the proposed SPGAT outperforms state-of-the-art methods on both synthetic and real-world datasets.
\end{enumerate}

\section{Related Work}
In this section, we review low-light image enhancement, Transformer for vision applications, and generative adversarial learning.
\subsection{Low-Light Image Enhancement}
As mentioned above, there are two categories of solutions to solve the LIE problem: 1) classical LIE techniques and 2) learning-based LIE solutions.

\subsubsection{Classical Low-Light Image Enhancement}
In \cite{histogram-equalization_tce,dsp_histogram,yun2011histogram}, histogram equalization (HE) and its variants are adopted to restrain the histograms of the enhanced images to satisfy some constraints.
Dong~\etal. in \cite{dehazing-lowlight} propose a dehazing-based LIE method.
In \cite{lowlight-SparseRepresentations}, Fotiadou~\etal. suggest a sparse representation model by approximating the low-light image patches in an appropriate dictionary to corresponding daytime images.
Motivated by Retinex theory, Yamasaki~\etal.~\cite{Denighting} separate the images into two components: reflectance and illumination, and then enhance the images using the reflectance component.
Although these classical methods can enhance images to some extent, they tend to produce artifacts on enhanced images or generate under enhancement results.
\begin{figure*}[!t]
\begin{center}
\begin{tabular}{ccccc}\includegraphics[width = 0.99\linewidth]{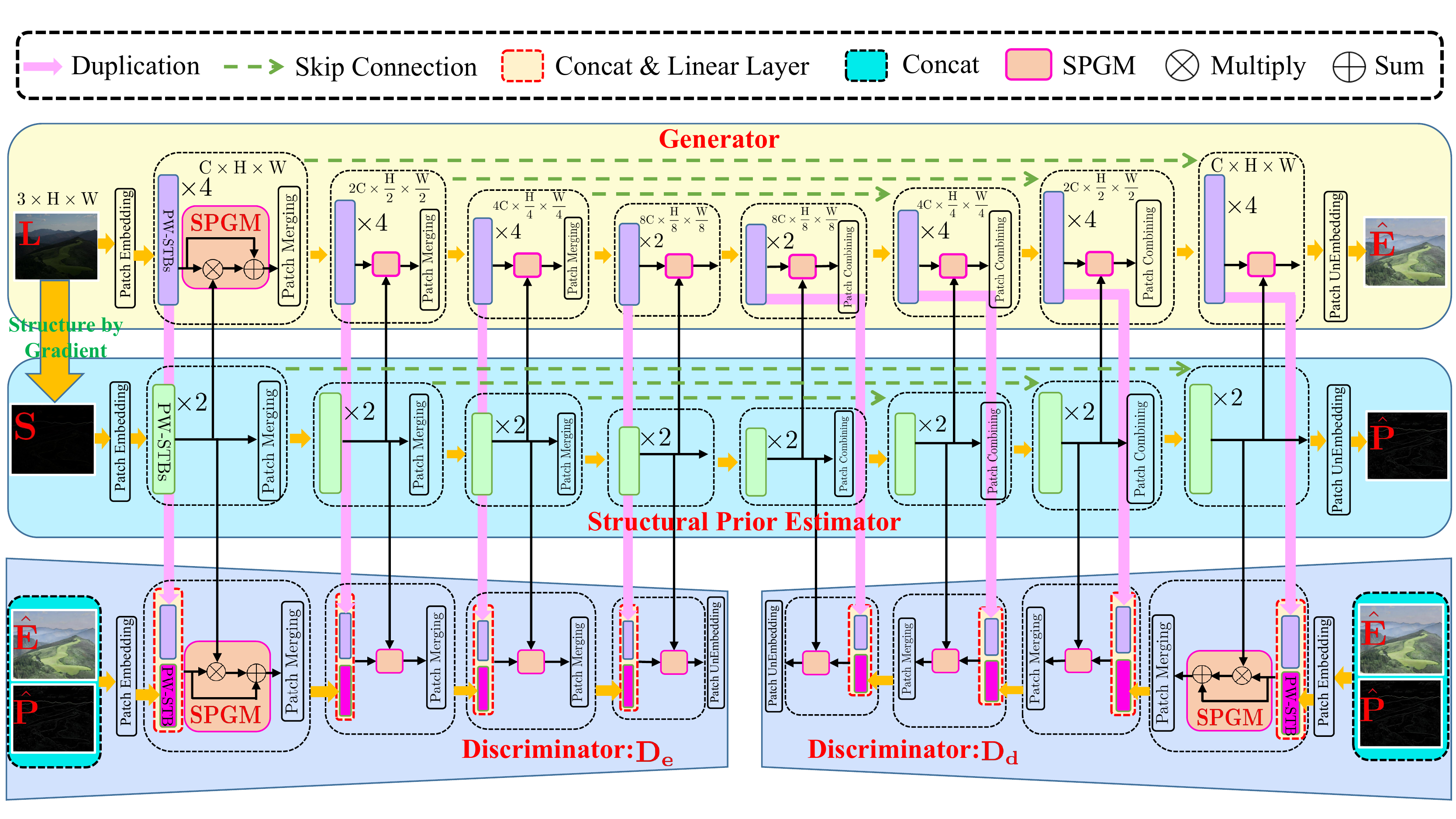}

\end{tabular}
\end{center}
\caption{Overview of the proposed Structural Prior guided Generative Adversarial Transformers (SPGAT).
The SPGAT contains one generator, two discriminators, and one structural prior estimator.
The generator is used to explore non-local information with the guidance of a structural prior estimator (SPE) for better clear image restoration.
We build the skip connections between the generator and discriminators so that the discriminators can better discriminate between real and fake features in the generator for generating more natural images.
Meanwhile, SPE is also utilized to guide the discriminators for better estimations.
The generator takes a low-light image $\mathrm{L}$ as the input and outputs an enhanced image $\hat{\mathrm{E}}$.
The SPE is input the corresponding structure $\mathrm{S}$ of $\mathrm{L}$ by gradient operation and outputs the estimated structural prior $\hat{\mathrm{P}}$.
Both the two discriminators take the concatenation of the enhanced image $\hat{\mathrm{E}}$ and estimated structural $\hat{\mathrm{P}}$ as the input, which is regarded as the fake label.
However, there is no real label feature.
To that aim, we generate real label features by inputting the normal-light image to the generator and its corresponding structure to SPE.
PW-STB illustrated in Fig.~\ref{fig: Parallel Windows-based Swin Transformer Block (PW-STB).} is the basic block of the generator, discriminators, and SPE.
}
\label{fig: Diagram of the proposed structural prior guided generative adversarial Transformers.}
\end{figure*}
\subsubsection{Learning-based Low-Light Image Enhancement}
With the great success of deep learning, LIE has achieved significant improvement.
Researchers have devoted their attention to designing varieties of deep networks to solve the LIE problem.
Lore~\etal.~\cite{pr-llnet} suggest a deep auto-encoder-based network to brighten low-light images by using a variant of the stacked-sparse denoising auto-encoder.
Some researchers also solve the LIE problem from the Retinex theory~\cite{Retinex_jei} in deep convolutional neural networks.
Wei~\etal.~\cite{retinexnet_wei_bmvc18} propose to simultaneously estimate the reflectance and illumination map from low-light images to produce enhanced results.
Zhang~\etal.~\cite{kind_zhang_mm19} also design a Retinex-based model that decomposes images into illumination for light adjustment and reflectance for degradation removal, which will facilitate to be better learned.
Wang~\etal.~\cite{Illumination-Estimation} study the intermediate illumination in a deep network to associate the input with expected enhancement result in a bilateral learning framework.
To better improve the details of the enhanced images, Xu~\etal.~\cite{xu_lowlight_cvpr20} decompose the low-light images into low- and high-frequency layers and the low-frequency layer is first used to restore image objects, and then high-frequency details are enhanced by the recovered image objects.
By reformulating LIE as an image-specific curve estimation problem, Guo~\etal.~\cite{zerodce_lowlight_guo} propose a zero-reference deep curve estimation model to enhance the images by implicitly measuring the enhancement quality.
To recover perceptual quality, Yang~\etal. in \cite{DRBN_yang_cvpr20} design a semi-supervised recursive band network with adversarial learning.
By combining the Retinex theory~\cite{Retinex_jei} and neural architecture search~\cite{nas} in a deep network, Liu~\etal.~\cite{RUAS_liu_cvpr21} design a Retinex-inspired unrolling model for LIE.
Several notable works \cite{dhn_ren_tip19,EEMEFN_aaai20_zhu} also use edge information in LIE.
In \cite{dhn_ren_tip19}, Ren~\etal. design a deep convolutional neural network (CNN) to enhance images equipped with spatially variant recurrent neural network to enhance details.
Zhu~\etal.~\cite{EEMEFN_aaai20_zhu} proposes a two-stage model with first multi-exposure fusion and then edge enhancement.

Although these works can achieve enhancement performance to some extent, all the above works are based on CNNs that do not effectively model the non-local information that may be useful for better clear image restoration.
To explore the non-local information, we introduce a new Transformer-based approach to solve the LIE problem.
We propose a new structural prior guided generative adversarial
Transformers and build the skip connections between the generator and discriminators with the guidance of the structural prior.
The proposed model adequately explores the global content by MLP architectures and the built adversarial learning with the skip connections simultaneously guided by the structural prior can effectively guide the discriminative process for facilitating better enhancement.
As we know, this is the first effort to explore the Transformer-based generative adversarial model with the skip connections between the generator and discriminators for low-light image enhancement.

%
\subsection{Transformer for Vision Applications}
Transformer is first proposed by~\cite{NIPS2017_Transformer} for natural language processing and then is extended to vision tasks (e.g., ViT~\cite{vit}).
Motivated by ViT, various Transformers have successfully developed for different vision tasks, e.g., segmentation~\cite{SegFormer}, detection~\cite{Liu_swintrans_ICCV2021,pvt_Wang_ICCV21}, and image restoration~\cite{ipt_Chen_CVPR2021,Liang_SwinIR_2021ICCV,wang2021uformer,ji2021u2,zamir2021restormer}.
%
%
However, directly using existing Transformers may not solve the LIE problem well as the LIE problem not only requires generating clear images with detailed structures but also needs to guarantee that the color of the restored image looks natural.
Hence, how to design an effective Transformer for LIE to produce more natural results with finer structures is worthy to studying.
%


\subsection{Generative Adversarial Learning}
Generative adversarial learning~\cite{gan,relativisticdiscriminator} achieves a significant process in image restoration tasks for realistic result generation, such as image dehazing~\cite{dehaze_cvpr18_gan,dcpdn,hardgan_eccv20_deng}, deraining~\cite{cgan_derain_zhang,pengxi_aaai2019}, deblurring~\cite{deblurgan_cvpr18,physicsgan_pan,deblurv2},
denoising~\cite{gan_denoising,tmi_denoise_gan},
super-resolution~\cite{SRGAN,esrgan,ranksrgan}, and also LIE~\cite{enlighten_tip_jiang}.
During adversarial learning, these methods usually use a standard discriminator to distinguish whether the generated image is fake or not.
However, this approach may lead to instability as gradients passing from the discriminator to the generator become uninformative when there is not enough overlap in the supports of the real and fake distributions~\cite{msg-gan_cvpr20}.
To solve this problem, Karnewar~\etal.~\cite{msg-gan_cvpr20} develop a multi-scale generative adversarial network that inputs the generated multi-scale images from intermediate layers to one discriminator.
However, we find that generated multi-scale images may not be accurate, which will affect the restoration quality.
More recently, several Transformer-based adversarial learning methods~\cite{icml2021gat,jiang2021transgan} are introduced to explore visual generative modeling.

Different from these methods, we propose a structural prior guided Transformer with adversarial training by building the skip connections between the generator and discriminators that directly transmit features from the generator to discriminators, and the learned features in discriminators are simultaneously guided by structural prior.
Such a design can help the generator generate more natural results with better image enhancement.

\section{Proposed Method}
We develop an effective Structural Prior guided Generative Adversarial Transformer (SPGAT) to better explore the structures and non-local information based on a Transformer in a GAN framework for low-light image enhancement.
Fig.~\ref{fig: Diagram of the proposed structural prior guided generative adversarial Transformers.} illustrates the proposed SPGAT, which consists of one Transformer generator, two Transformer discriminators, and one Transformer structural prior estimator, where the structural prior estimator is used to explore useful structures as the guidance of the generator and discriminators. In the following, we explain each module in detail.

\subsection{Network Architecture}
\subsubsection{Structural Prior Estimator}
Structural Prior Estimator (SPE) is a U-shaped Transformer with the proposed Parallel Windows-based Swin Transformer Block (PW-STB), which is input the structure $\mathrm{S}$ of the low-light image and estimates the structure $\mathrm{\hat{P}}$ of the normal-light one.
As the structure is easier than the image itself, SPE is able to better learn structural features to help guide not only the generator but also the discriminators for better image enhancement.
%
\subsubsection{Transformer Generator}
The generator is also a U-shaped Transformer architecture which has a similar architecture with SPE, which estimates the normal-light image $\mathrm{\hat{E}}$ from a given low-light image $\mathrm{L}$.
Given a $\mathrm{L}$, we use Patch Embedding to convert the image to tokens.
The converted image tokens are further extracted by a series of PW-STBs and Patch Merging to encode the tokens to deeper and deeper embedding space.
The decoder has a symmetric architecture with the encoder and the skip connection is utilized to receive the features at the symmetric blocks in the encoder and Patch Combining which is an inverse operation of Patch Merging is employed to shorten the dimension of embedding to shallower and shallower space.
Here, we use Patch UnEmbedding to convert the tokens to the image.
In the process of learning for generator, SPE also learns the structural features to guide the learning process of generator by Structure Prior Guide Module (SPGM):
\begin{equation}
\begin{array}{ll}
\mathrm{SPGM}(\mathbf{F}_{\mathrm{E}}) = \mathbf{F}_{\mathrm{P}} * \mathbf{F}_{\mathrm{E}} + \mathbf{F}_{\mathrm{E}},
\end{array}
\label{eq:spgm}
\end{equation}
where $\mathbf{F}_{\mathrm{E}}$ and $\mathbf{F}_{\mathrm{P}}$ respectively denote the embedding features from generator and SPE which will be introduced in the following.
$*$ and $+$ respectively refer to element-wise multiplication and summation.
Although the used SPGM is simpler, it does not require extra parameters and we will show that our SPGM is superior than the widely used concatenation fusion operation in Section~\ref{sec: Analysis on Basic Component}.
\begin{figure*}[!t]

\begin{center}
\begin{tabular}{ccccc}
\includegraphics[width = 0.97\linewidth]{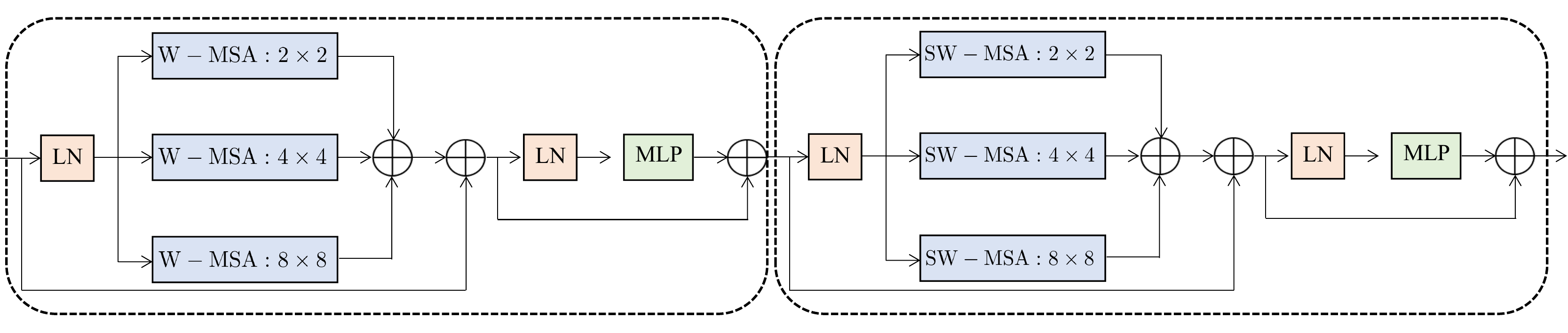}
\end{tabular}
\end{center}
\caption{Parallel Windows-based Swin Transformer Block (PW-STB).
'LN' denotes the Layer Normalization;
'W-MSA' and 'SW-MSA' respectively indicate the regular window-based multi-head self-attention modules and shifted window-based multi-head self-attention modules with different combinations of windows.
'MLP' refers to the multi-layer perceptron.
The PW-STB aggregates different level hierarchical features learned in different window sizes for better enhancing images.
}
\label{fig: Parallel Windows-based Swin Transformer Block (PW-STB).}
\end{figure*}
\subsubsection{Transformer Discriminators}
There are two Transformer discriminators, which respectively discriminate encoder and decoder embedding features.
We cascade the estimated image $\hat{\mathrm{E}}$ and structure $\hat{\mathrm{P}}$ to input to each discriminator and employ the Patch Embedding operation to convert the image to tokens.
We also utilize Patch Merging to encode the tokens to deeper and deeper dimensions and lastly we use Patch UnEmbedding to convert the tokens to 4-D tensor for computing the adversarial losses.
However, different from existing methods~\cite{cgan_derain_zhang,deblurgan_cvpr18,PhysicsGAN} that only input the image to the discriminator without considering the correlation between the generator and discriminator, we in this paper propose to build the skip connection between the generator and discriminators with the guidance of learned structural prior features by SPGM:
\begin{equation}
\begin{array}{ll}
\mathrm{SPGM}(\mathbf{F}_{\mathrm{D}}) = \mathbf{F}_{\mathrm{P}} * \mathcal{H}(\mathbf{F}_{\mathrm{D}}, \mathbf{F}_{\mathrm{E}})  + \mathcal{H}(\mathbf{F}_{\mathrm{D}}, \mathbf{F}_{\mathrm{E}}),
\end{array}
\label{eq:msgnn}
\end{equation}
where $\mathcal{H}$ denotes the skip connection between the generator and discriminator that is achieved by a concatenation operation followed with a Linear convolution layer.
$\mathbf{F}_{\mathrm{D}}$ denotes the features in discriminator.
%

\eqref{eq:msgnn} builds the connection between the features from generator (i.e., $\mathbf{F}_{\mathrm{E}}$) and the features in discriminator (i.e., $\mathbf{F}_{\mathrm{D}}$) and the connection is simultaneously guided by learned immediate features in structural prior estimator (i.e., $\mathbf{F}_{\mathrm{P}}$).
Such a design not only overcomes the uninformative gradients passing from the discriminator to the generator~\cite{msg-gan_cvpr20} but also simplifies the process for generating images in \cite{msg-gan_cvpr20} and avoids generating abnormal images.


\subsection{Parallel Windows-Based Swin Transformer Block}
Swin Transformer block is proposed in \cite{Liu_swintrans_ICCV2021}, where it conducts multi-head self-attention in one regular/shifted window.
Different from \cite{Liu_swintrans_ICCV2021} that only exploits the self-attention in one window in the Transformer layer that does not consider the different level hierarchical features from different window sizes, we in this paper develop a Parallel Windows-based Swin Transformer Block (PW-STB) that use multi parallel windows to replace the original one window in \cite{Liu_swintrans_ICCV2021} to obtain different level hierarchical features in different window sizes.
And the learned features in each window are added to obtain the aggregated features in the window-level self-attention.
Fig.~\ref{fig: Parallel Windows-based Swin Transformer Block (PW-STB).} illustrates the layout of PW-STB.


With the parallel style, one PW-STB is computed by:
\begin{equation}
\begin{array}{ll}
\hat{\mathrm{z}}^{l} =& \mathop{\mathrm{W}\text{-}\mathrm{MSA}}\limits_{2\times 2}\big(\mathrm{LN}(z^{l-1})\big) + \mathop{\mathrm{W}\text{-}\mathrm{MSA}}\limits_{4\times 4} \big(\mathrm{LN}(z^{l-1})\big)+ \\
&
\mathop{\mathrm{W}\text{-}\mathrm{MSA}}\limits_{8\times 8}\big(\mathrm{LN}(z^{l-1})\big)
+ z^{l-1},
\\
\\
\mathrm{z}^{l} = & \mathrm{MLP}\big(\mathrm{LN}(\hat{\mathrm{z}}^{l})\big) + \hat{\mathrm{z}}^{l},
\\
\\
\hat{\mathrm{z}}^{l+1}  =& \mathop{\mathrm{SW}\text{-}\mathrm{MSA}}\limits_{2\times 2}\big(\mathrm{LN}(z^{l})\big) + \mathop{\mathrm{SW}\text{-}\mathrm{MSA}}\limits_{4\times 4} \big(\mathrm{LN}(z^{l})\big) +
\\
& \mathop{\mathrm{SW}\text{-}\mathrm{MSA}}\limits_{8\times 8}\big(\mathrm{LN}(z^{l})\big) + z^{l},
\\
\\
\mathrm{z}^{l+1} = &\mathrm{MLP}\big(\mathrm{LN}(\hat{\mathrm{z}}^{l+1})\big) + \hat{\mathrm{z}}^{l+1},
\end{array}
\label{eq:pw-stb}
\end{equation}
where $\hat{\mathrm{z}}^{l}$ and $\mathrm{z}^{l}$ respectively denote the output features of the (S)W\text{-}MSA module and the MLP module for block $l$.
$\mathop{\mathrm{W}\text{-}\mathrm{MSA}}\limits_{k\times k}$ and $\mathop{\mathrm{SW}\text{-}\mathrm{MSA}}\limits_{k\times k}$ indicate the the regular window and shifted window with $k \times k$ window size, respectively.

\subsection{Loss Function}\label{sec: Loss function}
To train the network, we utilize four objective loss functions including image pixel reconstruction loss ($\mathcal{L}_{i}$), structure reconstruction loss ($\mathcal{L}_{s}$), and two adversarial losses ($\mathcal{L}^{e}_{a}$ and $\mathcal{L}^{d}_{a}$):
\begin{equation}
\mathcal{L} = \mathcal{L}_{i} + \alpha \mathcal{L}_{s}  + \beta (\mathcal{L}^{e}_{a} + \mathcal{L}^{d}_{a}),
\label{eq:total}
\end{equation}
where $\alpha$ and $\beta$ denote the hyper-parameters. In the following, we explain each term in details.
\\
\textbf{Image pixel reconstruction loss.}
The SSIM-based loss has been applied to image deraining~\cite{derain_prenet_Ren_2019_CVPR,dcsfn-wang-mm20} and achieves better performance, we use it as the pixel reconstruction loss:
\begin{equation}
\mathcal{L}_{i} = 1-\mathrm{SSIM}\big(\hat{\mathrm{E}}, \mathrm{E}\big),
\label{eq:losspixel}
\end{equation}
where $\hat{\mathrm{E}}$ and $\mathrm{E}$ denote the estimated image and corresponding ground-truth.
\\
\textbf{Structure reconstruction loss.}
We use $L_{1}$ loss to measure the reconstruction error between estimated structural prior $\hat{\mathrm{P}}$ and corresponding ground-truth structural prior $\mathrm{P}$:
\begin{equation}
\mathcal{L}_{s} = ||\hat{\mathrm{P}}- \mathrm{P}||_{1}.
\label{eq:losspixel}
\end{equation}
\\
\textbf{Adversarial losses.} To better help generator generate more natural results, we develop two discriminators $\mathrm{D}_{\mathrm{e}}$ and $\mathrm{D}_{\mathrm{d}}$ by building the skip connection between the generator and discriminators to respectively transmit the encoder and decoder features in the generator to the discriminator $\mathrm{D}_{\mathrm{e}}$ and the discriminator $\mathrm{D}_{\mathrm{d}}$ so that the two discriminators can better discriminate between real and fake features.
The two adversarial losses about the two discriminators are defined as:
\begin{equation}
\begin{array}{ll}
\mathcal{L}^{e}_{a} = &-\mathbb{E}_{\mathbf{X}^{e}}\Big[\log\big(1-D_{ra}(\mathbf{X}^{e};\mathbf{Y}^{e})\big)\Big]-
\\
&
\mathbb{E}_{\mathbf{Y}^{e}}\Big[\log\big(D_{ra}(\mathbf{Y}^{e};\mathbf{X}^{e})\big)\Big],
\end{array}
\label{eq:lossencoder}
\end{equation}
and
\begin{equation}
\begin{array}{ll}
\mathcal{L}^{d}_{a} = &-\mathbb{E}_{\mathbf{X}^{d}}\Big[\log\big(1-D_{ra}(\mathbf{X}^{d};\mathbf{Y}^{d})\big)\Big]-
\\
&
\mathbb{E}_{\mathbf{Y}^{d}}\Big[\log\big(D_{ra}(\mathbf{Y}^{d};\mathbf{X}^{d})\big)\Big],
\end{array}
\label{eq:lossdecoder}
\end{equation}
where $\mathbf{Y}^{e}$ ($\mathbf{Y}^{d}$) denotes the combination among $\hat{\mathrm{E}}$, $\hat{\mathrm{P}}$, encoder (decoder) features in the generator, and the corresponding guidance structure features in the SPE, which is regarded as the fake label.
$\mathbf{X}^{e}$ ($\mathbf{X}^{d}$) is the corresponding real label.
$D_{ra}(\mathbf{U};\mathbf{V}) = \text{sigmoid}\Big(\mathrm{D}_{\mathrm{e}}(\mathbf{U})-\mathbb{E}_{\mathbf{Q}}\big[\mathrm{D}_{\mathrm{e}}(\mathbf{V})\big]\Big)$ for \eqref{eq:lossencoder}; $D_{ra}(\mathbf{U};\mathbf{V}) = \text{sigmoid}\Big(\mathrm{D}_{\mathrm{d}}(\mathbf{U})-\mathbb{E}_{\mathbf{Q}}\big[\mathrm{D}_{\mathrm{d}}(\mathbf{V})\big]\Big)$ for \eqref{eq:lossdecoder}.
However, there is not real label feature.
\textbf{\textit{To this end, we generate the real label features by inputting the normal-light image to generator and its structure to SPE.}}

\section{Experiments}
In this section, we compare our method with recent state-of-the-art methods, including Retinex~\cite{retinexnet_wei_bmvc18}, KinD~\cite{kind_zhang_mm19}, Enlighten~\cite{enlighten_tip_jiang}, RRDNet~\cite{RRRDNet_zhu_icme2020}, DeepUPE~\cite{deepupe_cvpr19}, DRBN~\cite{DRBN_yang_cvpr20}, FIDE~\cite{fide}, Zero-DCE~\cite{zerodce_lowlight_guo},  Zero-DCE++~\cite{zerodce_lowlight_pami}, RUAS~\cite{RUAS_liu_cvpr21}.
Extensive analysis is also conducted to verify the effectiveness of the proposed approach.

%

\subsection{Network and Implementation Details}
%
There are 4, 4, 4, and 2 PW-STBs in the encoder layer of the generator and 2, 2, 2, and 2 PW-STBs in the encoder layer of SPE, and their decoders respectively have a symmetric number of PW-STBs.
For the discriminators, there is 1 PW-STB in each layer.
For the self-attention in the PW-STB of each layer, the number of heads is 4.
$\mathrm{C}$ in Fig.~\ref{fig: Diagram of the proposed structural prior guided generative adversarial Transformers.} is set as 32.
We randomly crop $128 \times 128$ patch as input and the batch size is $2$.
We use ADAM~\cite{adam} to train the model.
The initial learning rate is 0.0001, which will be divided by $2$ every 30 epochs, and the model training terminates after $150$ epochs.
$\alpha$ is 0.1 and $\beta$ is 0.001.
The updated radio $r$ between the training generator and discriminator is 5.
Our model is trained on one NVIDIA 3090 GPU based on the Pytorch framework.
The source code will be available if the paper can be accepted.

\begin{table*}[!t]
\centering
\caption{Comparisons with baselines on the LOL dataset.
The best results are marked in $\textbf{blod}$.
$\uparrow$ refers to that higher is better.
All the methods are retrained and results are recomputed in RGB space by our PSNR and SSIM codes.
$^{*}$ denotes the authors only provide testing codes so we can only use the released models for testing.
}
\scalebox{0.95}{
\begin{tabular}{lccccccccccccc}
\toprule
\multirow{2}{*}{Dataset}        &  \multirow{2}{*}{Metrics}    & Retinex & KinD   & Enlighten & RRDNet& DeepUPE$^{*}$&DRBN & FIDE$^{*}$& Zero-DCE  & Zero-DCE++ & RUAS   & SPGAT   \\
        &     & \cite{retinexnet_wei_bmvc18} & \cite{kind_zhang_mm19}   & \cite{enlighten_tip_jiang} & \cite{RRRDNet_zhu_icme2020} &\cite{deepupe_cvpr19}&\cite{DRBN_yang_cvpr20}& \cite{fide}& \cite{zerodce_lowlight_guo}  & \cite{zerodce_lowlight_pami} & \cite{RUAS_liu_cvpr21}   & (Ours)   \\

\toprule
\multirow{2}{*}{LOL}         & PSNR $\uparrow$ & 17.02      & 17.94  & 17.95       & 12.06&12.71&18.79&18.34&  16.04      & 14.75      & 16.34  & $\textbf{19.80}$  \\ 
                             & SSIM $\uparrow$ & 0.4341     & 0.7804 & 0.6597       & 0.4680&0.4566&0.8014&0.8004 & 0.5240     & 0.5257     & 0.5044 & $\textbf{0.8234}$ \\
\bottomrule
\end{tabular}}
\label{tab:syn-Results-sota}
\end{table*}

\begin{figure*}[!t]
\begin{center}
\begin{tabular}{ccccc}
\includegraphics[width = 0.243\linewidth]{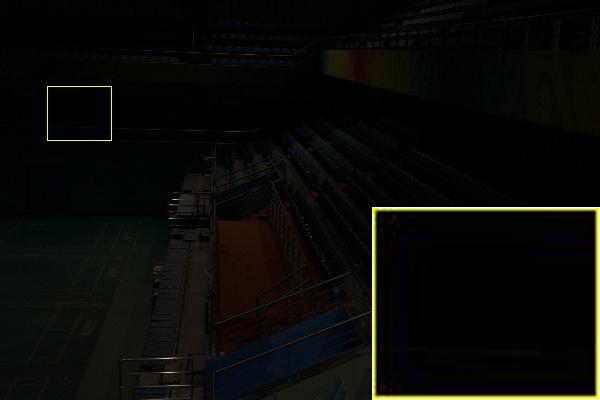}&\hspace{-4.5mm}
\includegraphics[width = 0.243\linewidth]{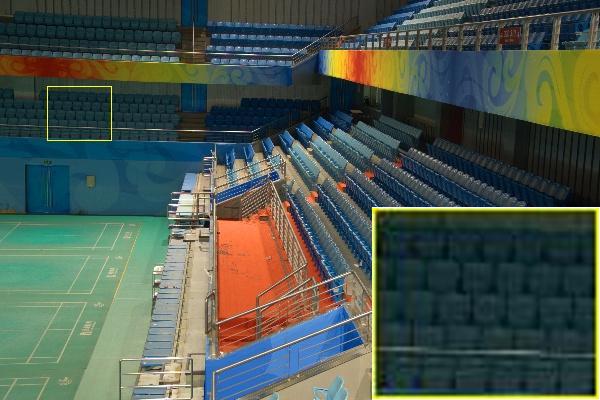}&\hspace{-4.5mm}
\includegraphics[width = 0.243\linewidth]{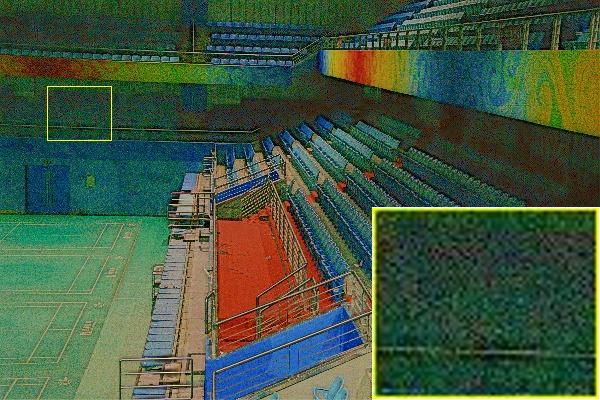}&\hspace{-4.5mm}
\includegraphics[width = 0.243\linewidth]{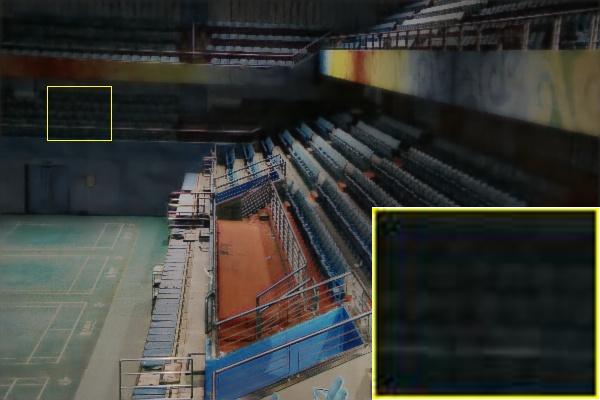}

\\
(a) Low-light &\hspace{-4mm} (b)  GT  &\hspace{-4mm} (c)  Retinex~\cite{retinexnet_wei_bmvc18}&\hspace{-4mm} (d) KinD~\cite{kind_zhang_mm19}
\\
\includegraphics[width = 0.243\linewidth]{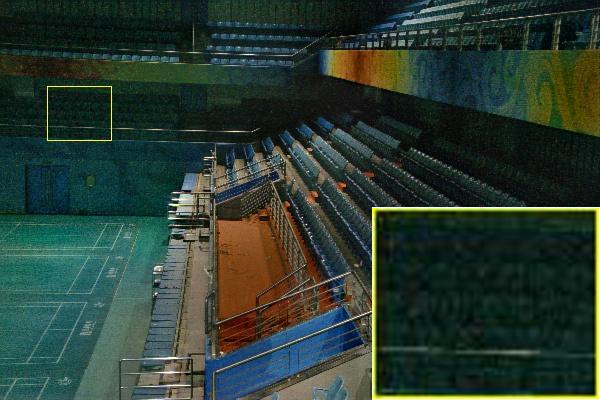}&\hspace{-4mm}
\includegraphics[width = 0.243\linewidth]{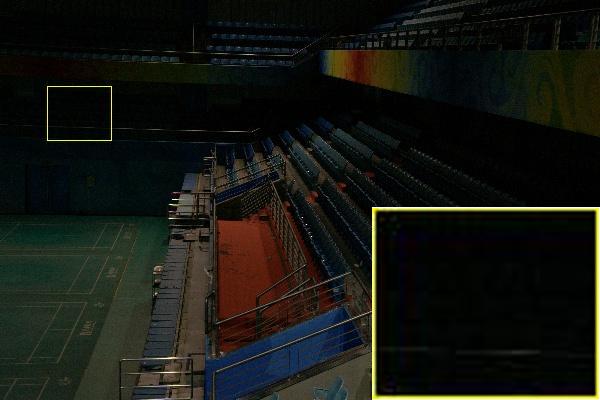}&\hspace{-4.5mm}
\includegraphics[width = 0.243\linewidth]{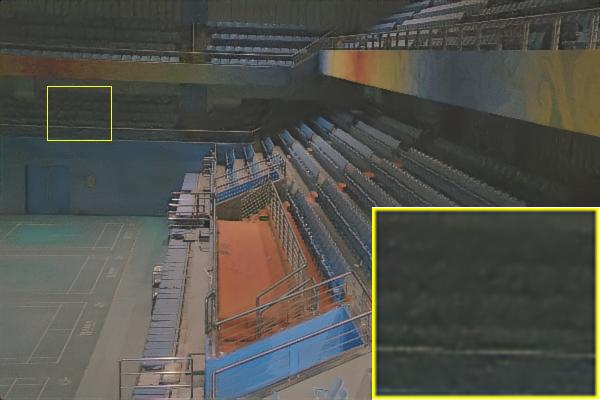}&\hspace{-4.5mm}
\includegraphics[width = 0.243\linewidth]{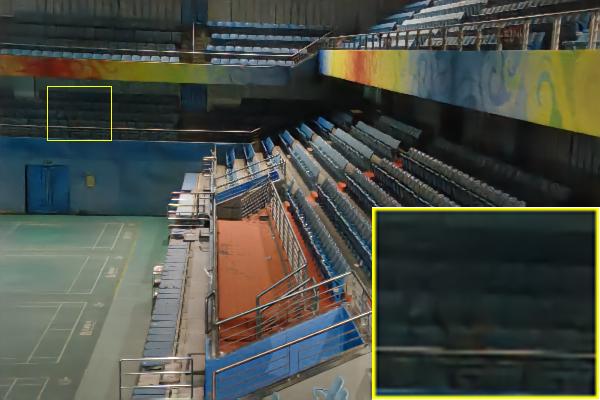}

\\
 (e) Enlighten~\cite{enlighten_tip_jiang}&\hspace{-4mm} (f) RRDNet~\cite{RRRDNet_zhu_icme2020}&\hspace{-4mm} (g)  DeepUPE~\cite{deepupe_cvpr19}&\hspace{-4mm}(h) DRBN~\cite{DRBN_yang_cvpr20}
\\
\includegraphics[width = 0.243\linewidth]{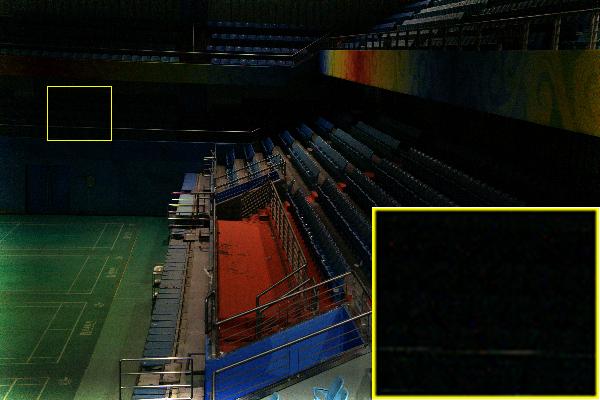}&\hspace{-4.5mm}
\includegraphics[width = 0.243\linewidth]{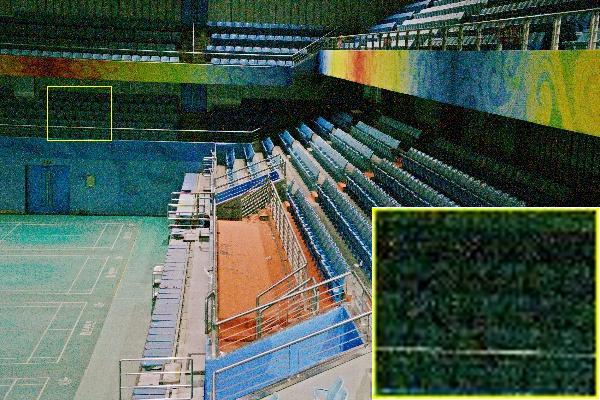}&\hspace{-4.5mm}
\includegraphics[width = 0.243\linewidth]{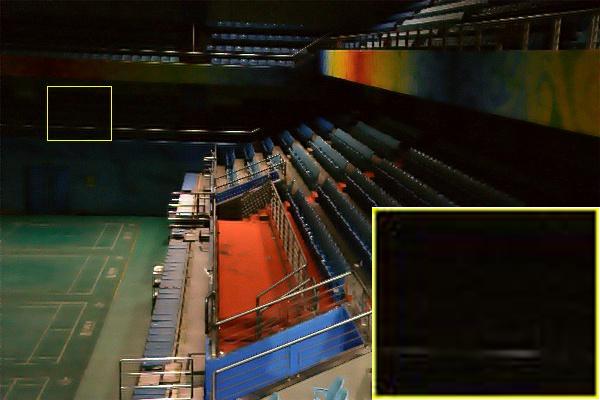}&\hspace{-4.5mm}
\includegraphics[width = 0.243\linewidth]{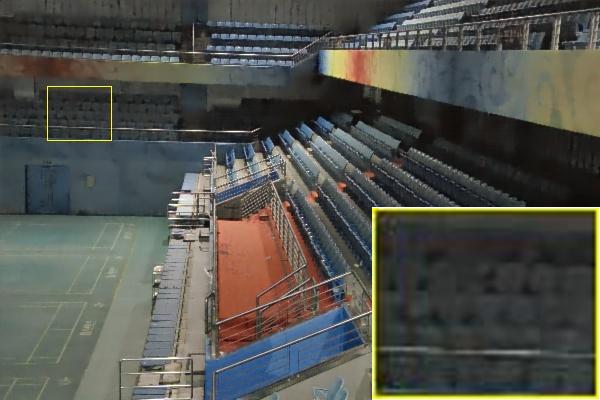}

\\
(i)  FIDE~\cite{fide}&\hspace{-4mm} (j) Zero-DCE~\cite{zerodce_lowlight_guo}&\hspace{-4mm} (k)  RUAS~\cite{RUAS_liu_cvpr21}&\hspace{-4mm} (l) SPGAT
\\

\end{tabular}
\end{center}
\caption{Comparisons with state-of-the-art approaches on the LOL dataset.
The proposed SPGAT is able to recover a clearer result with better structure.
}
\label{fig: Comparisons on the LOL dataset3.}
\end{figure*}

\begin{table*}[!t]
\centering
\caption{Comparisons with baselines on the Brightening dataset in terms of PSNR and SSIM.
}
\scalebox{0.93}{
\begin{tabular}{lccccccccccccc}
\toprule
                    \multirow{2}{*}{Dataset}        &  \multirow{2}{*}{Metrics}    & Retinex & KinD   & Enlighten & RRDNet& DeepUPE$^{*}$&DRBN & FIDE$^{*}$& Zero-DCE  & Zero-DCE++ & RUAS   & SPGAT   \\
        &     & \cite{retinexnet_wei_bmvc18} & \cite{kind_zhang_mm19}   & \cite{enlighten_tip_jiang} & \cite{RRRDNet_zhu_icme2020} &\cite{deepupe_cvpr19}&\cite{DRBN_yang_cvpr20}& \cite{fide}& \cite{zerodce_lowlight_guo}  & \cite{zerodce_lowlight_pami} & \cite{RUAS_liu_cvpr21}   & (Ours)   \\

\toprule

\multirow{2}{*}{Brightening} & PSNR$\uparrow$ & 17.13      &18.82  & 16.48        & 14.83&13.81&18.19&15.34  & 16.85     & 15.19      & 13.70  & \textbf{22.19}  \\ 
                             & SSIM$\uparrow$ & 0.7628     & 0.8436 & 0.7760       & 0.6540&0.6129&0.8662&0.6998 & 0.8118    & 0.7926     & 0.5830 & \textbf{0.9136} \\
\bottomrule
\end{tabular}
}
\label{tab:Comparisons with baselines on the Brightening dataset.}
\end{table*}

\begin{figure*}[!t]
\begin{center}
\begin{tabular}{ccccc}
\includegraphics[width = 0.243\linewidth]{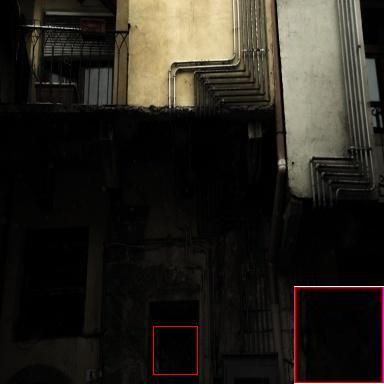}&\hspace{-4.5mm}
\includegraphics[width = 0.243\linewidth]{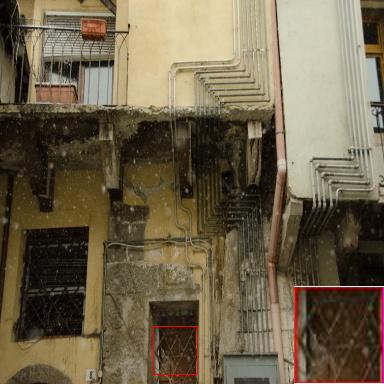}&\hspace{-4.5mm}
\includegraphics[width = 0.243\linewidth]{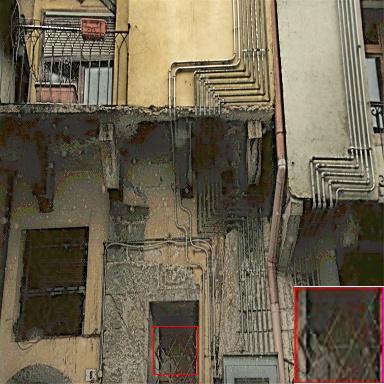}&\hspace{-4.5mm}
\includegraphics[width = 0.243\linewidth]{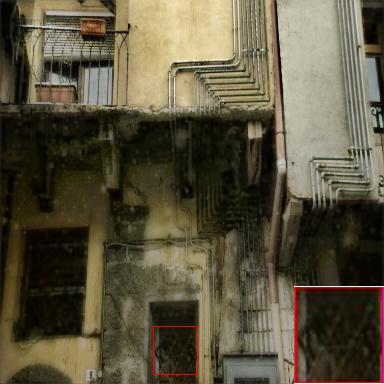}

\\
(a) Low-light &\hspace{-4mm} (b)  GT  &\hspace{-4mm} (c)  Retinex~\cite{retinexnet_wei_bmvc18}&\hspace{-4mm} (d) KinD~\cite{kind_zhang_mm19}
\\
\includegraphics[width = 0.243\linewidth]{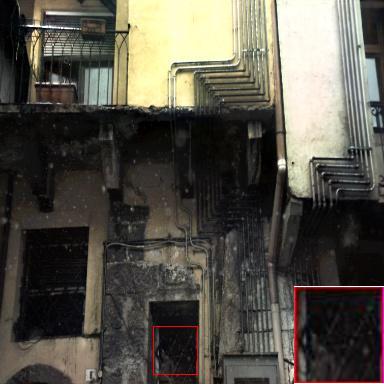}&\hspace{-4.5mm}
\includegraphics[width = 0.243\linewidth]{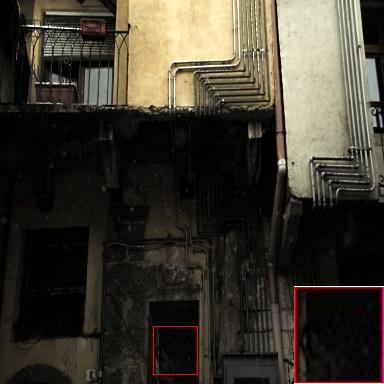}&\hspace{-4.5mm}
\includegraphics[width = 0.243\linewidth]{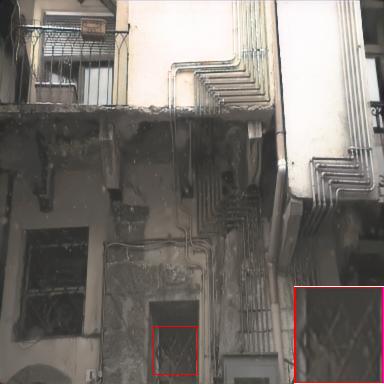}&\hspace{-4.5mm}
\includegraphics[width = 0.243\linewidth]{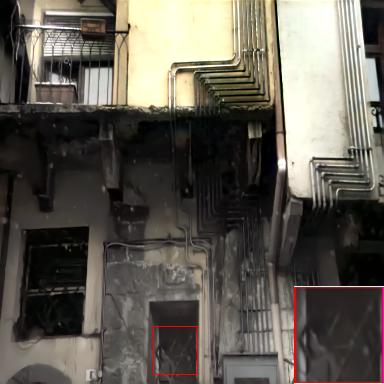}&\hspace{-4mm}

\\
 (e) Enlighten~\cite{enlighten_tip_jiang}&\hspace{-4mm} (f) RRDNet~\cite{RRRDNet_zhu_icme2020}&\hspace{-4mm} (g)  DeepUPE~\cite{deepupe_cvpr19}&\hspace{-4mm}(h) DRBN~\cite{DRBN_yang_cvpr20}
\\
\includegraphics[width = 0.243\linewidth]{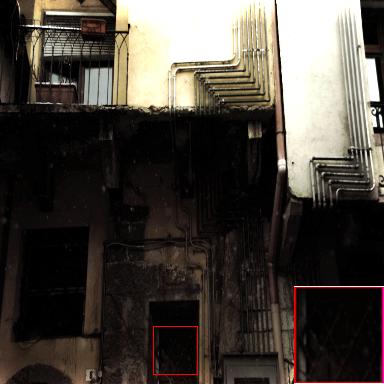}&\hspace{-4.5mm}
\includegraphics[width = 0.243\linewidth]{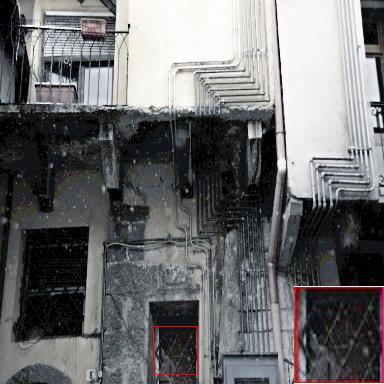}&\hspace{-4.5mm}
\includegraphics[width = 0.243\linewidth]{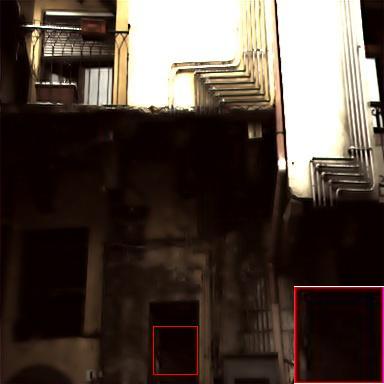}&\hspace{-4mm}
\includegraphics[width = 0.243\linewidth]{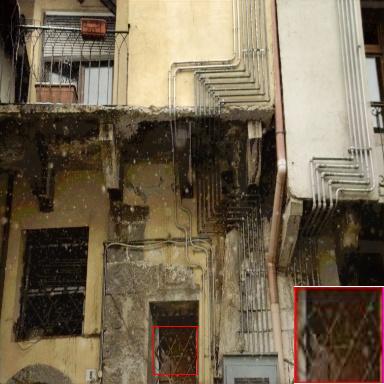}

\\
(i)  FIDE~\cite{fide}&\hspace{-4mm} (j) Zero-DCE~\cite{zerodce_lowlight_guo}&\hspace{-4mm} (k)  RUAS~\cite{RUAS_liu_cvpr21}&\hspace{-4mm} (l) SPGAT
\\

\end{tabular}
\end{center}
\caption{Comparisons with state-of-the-art approaches on the Brightening dataset.
The proposed SPGAT is able to recover a clearer result with finer texture.
}
\label{fig: Comparisons on the bri dataset2.}
\end{figure*}

\begin{figure*}[!t]
\begin{center}
\begin{tabular}{ccccc}
\includegraphics[width = 0.243\linewidth]{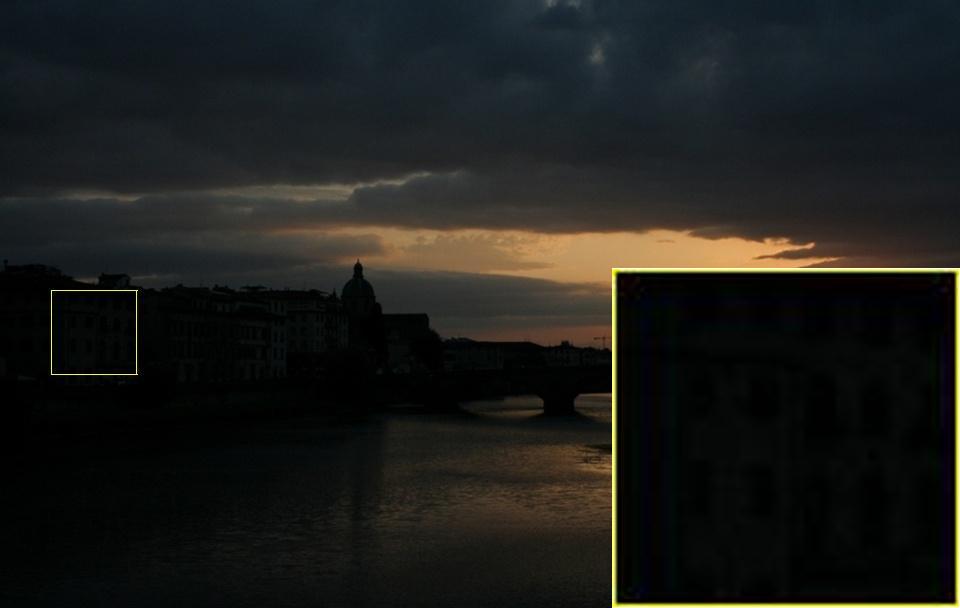}&\hspace{-4.5mm}
\includegraphics[width = 0.243\linewidth]{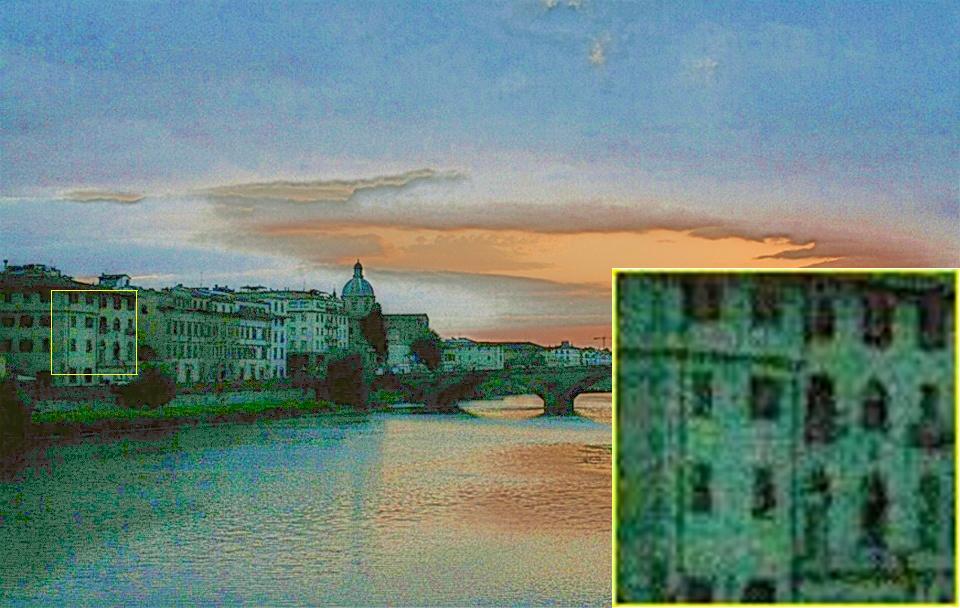}&\hspace{-4.5mm}
\includegraphics[width = 0.243\linewidth]{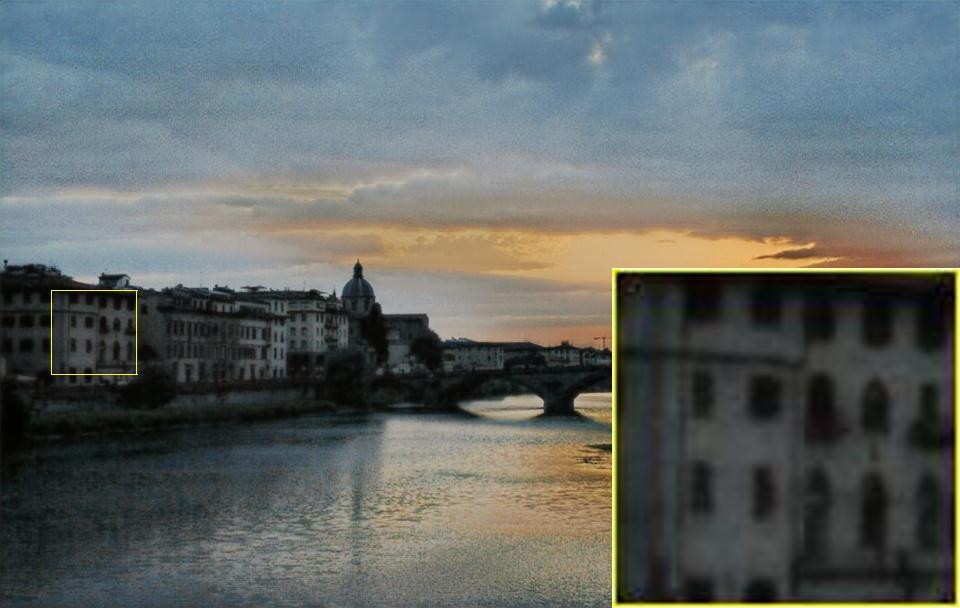}&\hspace{-4.5mm}
\includegraphics[width = 0.243\linewidth]{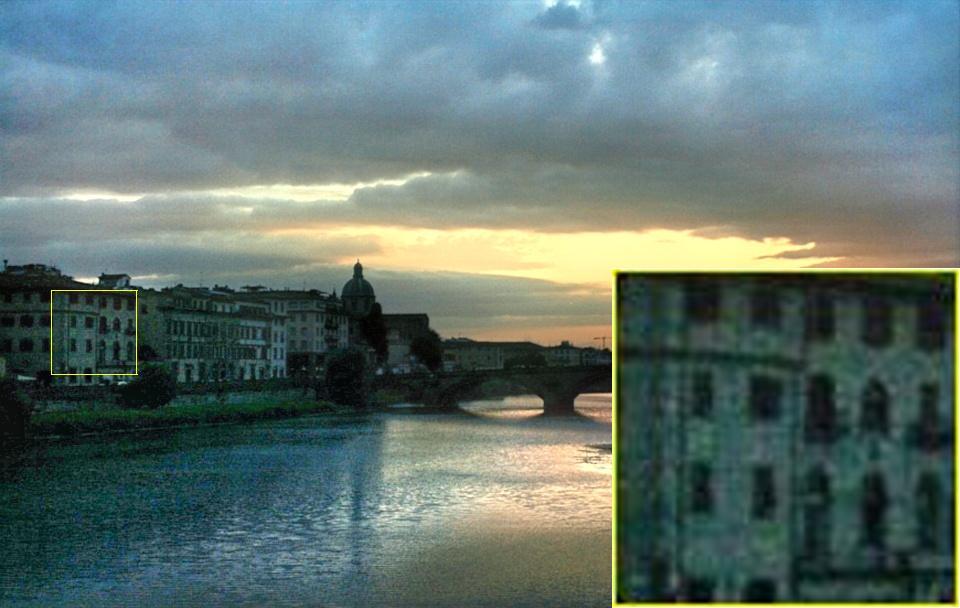}
\\
(a) Low-light &\hspace{-4mm} (b) Retinex~\cite{retinexnet_wei_bmvc18} &\hspace{-4mm} (c) KinD~\cite{kind_zhang_mm19} &\hspace{-4mm} (d) Enlighten~\cite{enlighten_tip_jiang}
\\
\includegraphics[width = 0.243\linewidth]{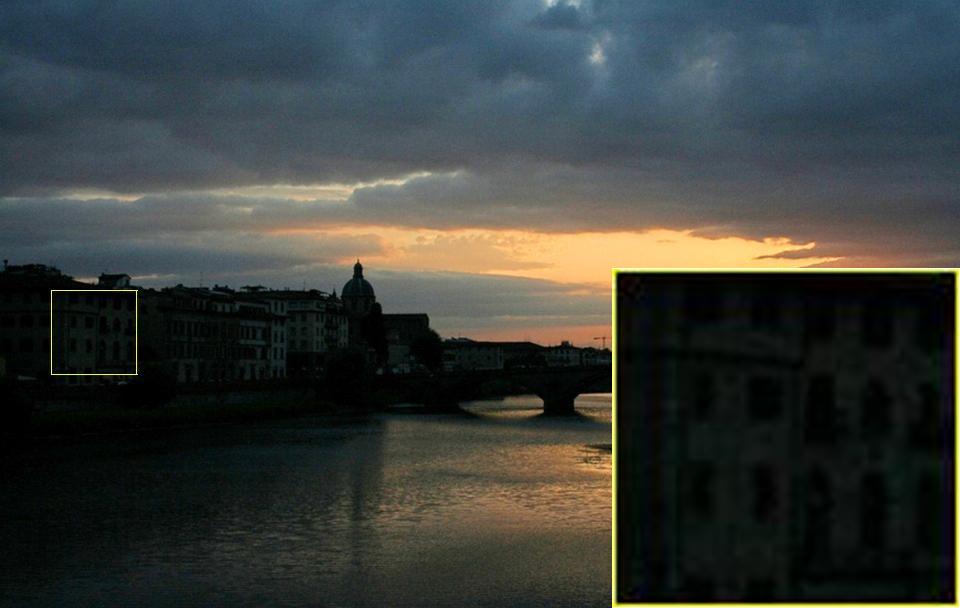}&\hspace{-4.5mm}
\includegraphics[width = 0.243\linewidth]{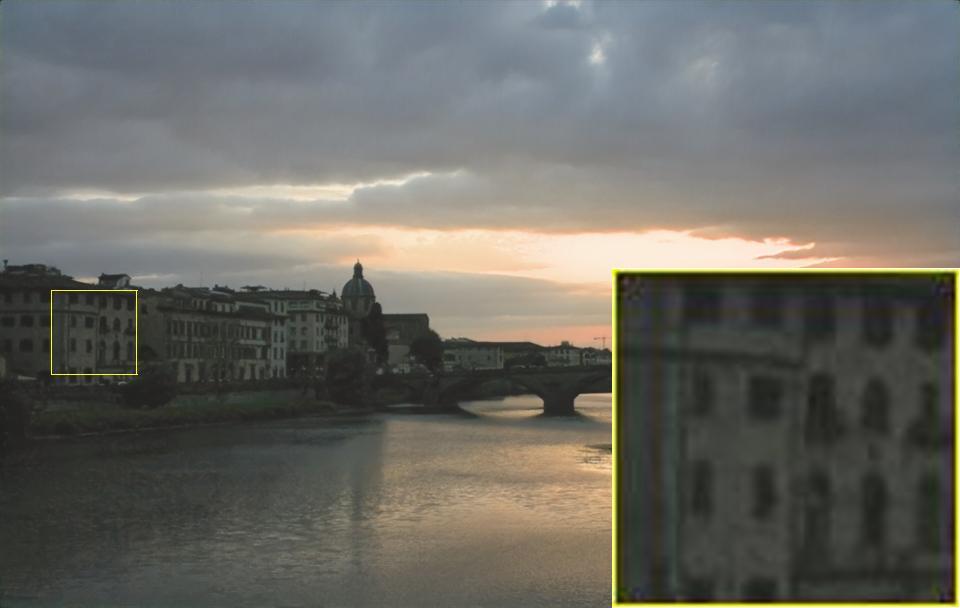}&\hspace{-4.5mm}
\includegraphics[width = 0.243\linewidth]{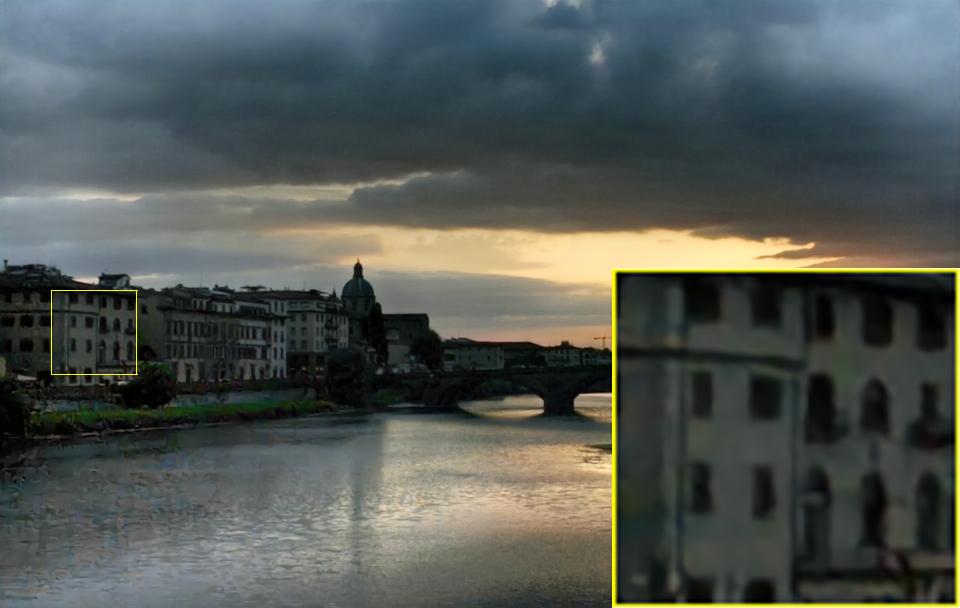}&\hspace{-4.5mm}
\includegraphics[width = 0.243\linewidth]{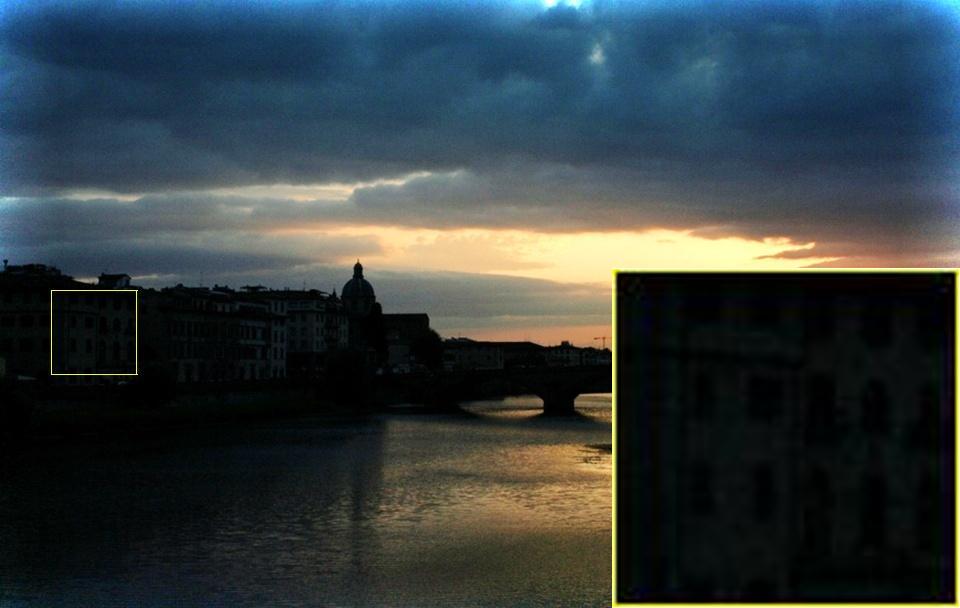}
\\
(e) RRDNet~\cite{RRRDNet_zhu_icme2020}&\hspace{-4mm} (f) DeepUPE~\cite{deepupe_cvpr19}&\hspace{-4mm} (g) DRBN~\cite{DRBN_yang_cvpr20} &\hspace{-4mm}(h) FIDE~\cite{fide}
\\
\includegraphics[width = 0.243\linewidth]{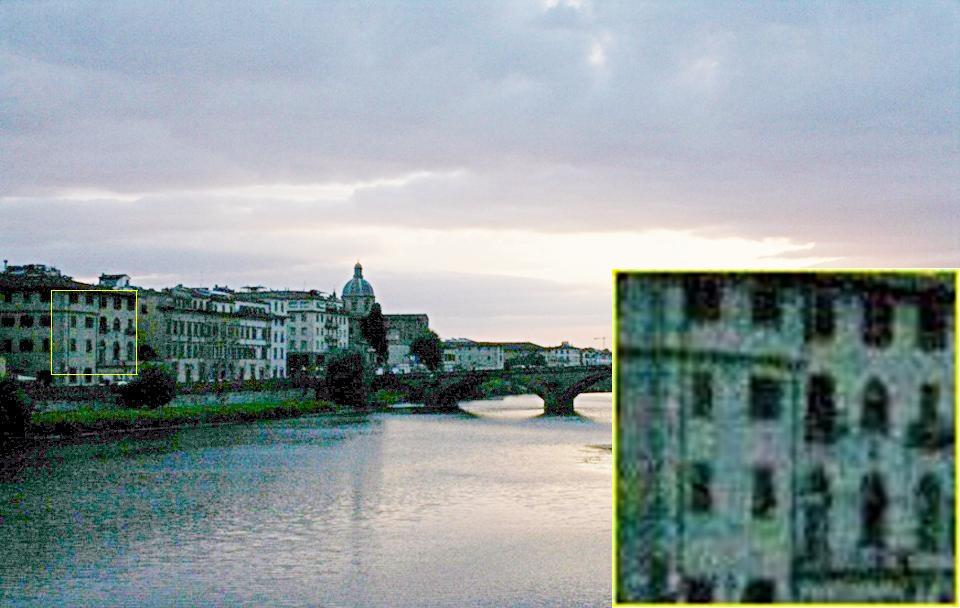}&\hspace{-4.5mm}
\includegraphics[width = 0.243\linewidth]{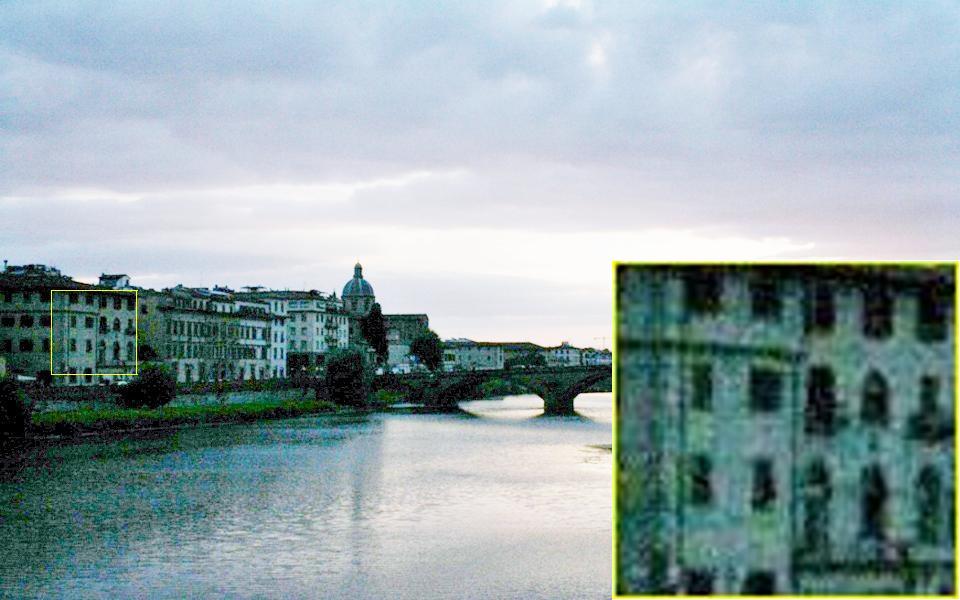}&\hspace{-4.5mm}
\includegraphics[width = 0.243\linewidth]{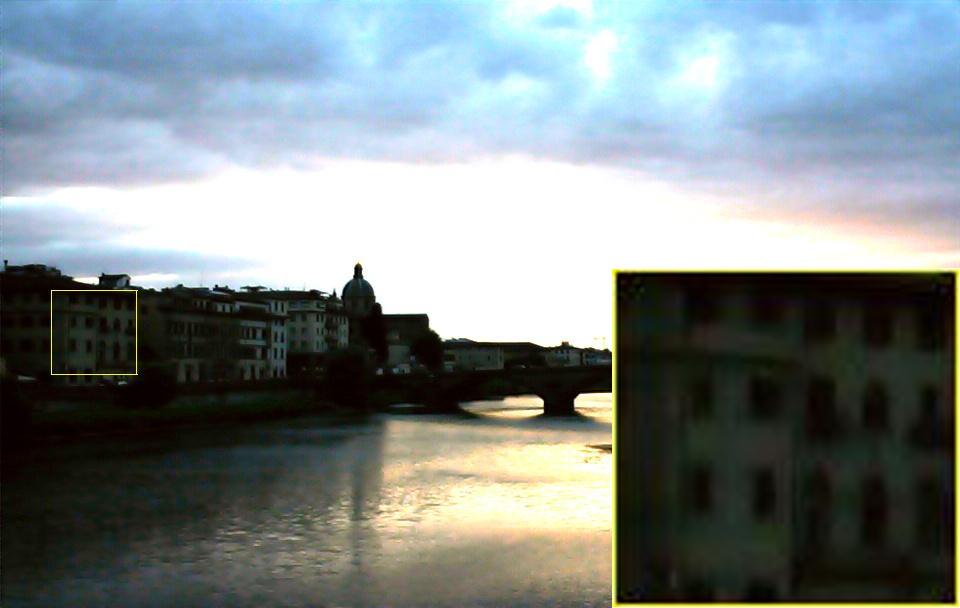}&\hspace{-4.5mm}
\includegraphics[width = 0.243\linewidth]{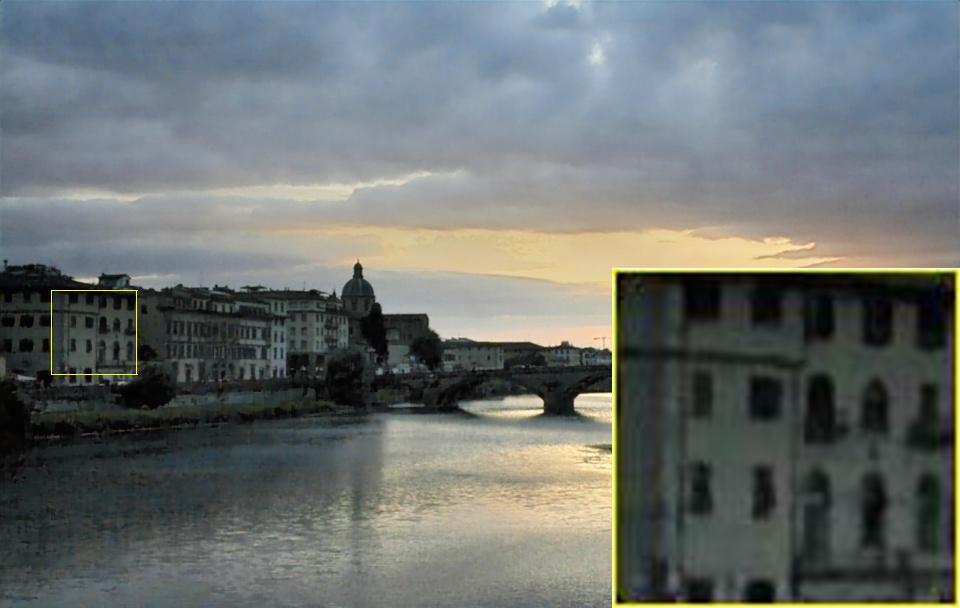}
\\
(i) Zero-DCE~\cite{zerodce_lowlight_guo} &\hspace{-4mm} (j) Zero-DCE++~\cite{zerodce_lowlight_pami}&\hspace{-4mm} (k) RUAS~\cite{RUAS_liu_cvpr21}&\hspace{-4mm} (l) SPGAT
\\

\end{tabular}
\end{center}
\caption{Comparisons with state-of-the-art methods on real-world images in the DICM dataset~\cite{dicm}.
The proposed SPGAT is able to generate a more natural result.
}
\label{fig: Comparisons on DICM images.}
\end{figure*}

\begin{figure*}[!t]
\begin{center}
\begin{tabular}{ccccc}
\includegraphics[width = 0.243\linewidth]{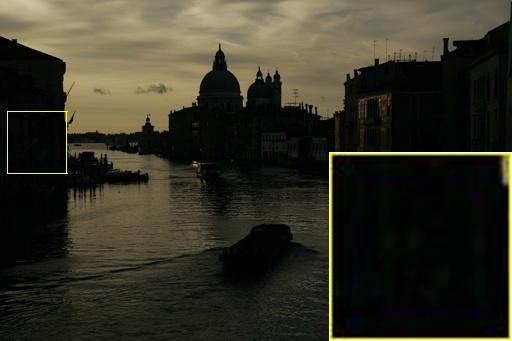}&\hspace{-4.5mm}
\includegraphics[width = 0.243\linewidth]{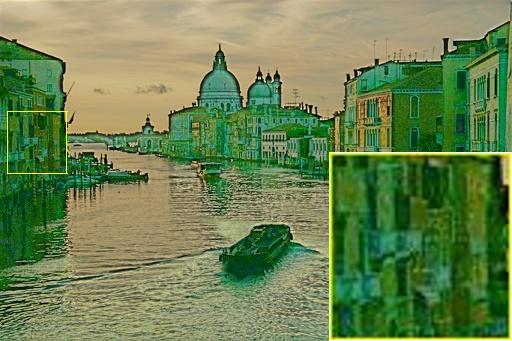}&\hspace{-4.5mm}
\includegraphics[width = 0.243\linewidth]{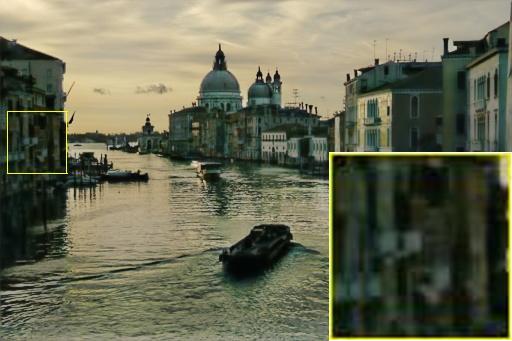}&\hspace{-4.5mm}
\includegraphics[width = 0.243\linewidth]{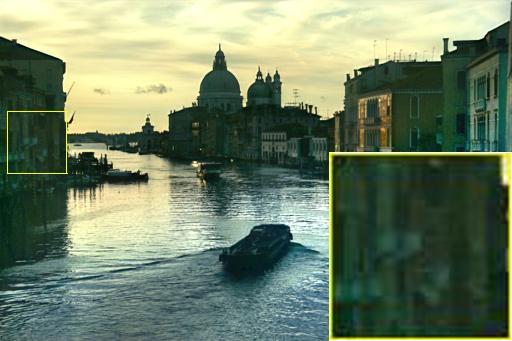}
\\
(a) Low-light &\hspace{-4mm} (b) Retinex~\cite{retinexnet_wei_bmvc18} &\hspace{-4mm} (c) KinD~\cite{kind_zhang_mm19} &\hspace{-4mm} (d) Enlighten~\cite{enlighten_tip_jiang}
\\
\includegraphics[width = 0.243\linewidth]{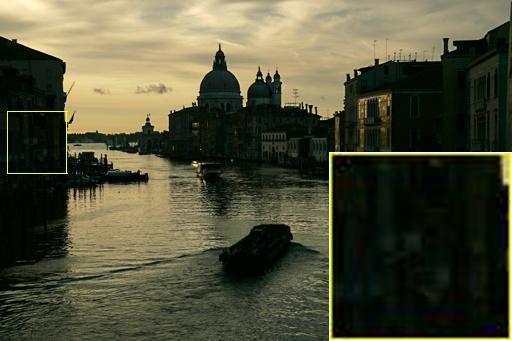}&\hspace{-4.5mm}
\includegraphics[width = 0.243\linewidth]{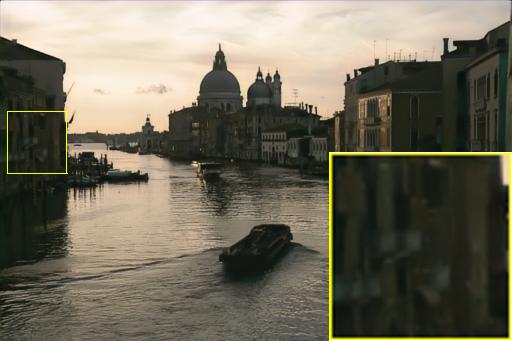}&\hspace{-4.5mm}
\includegraphics[width = 0.243\linewidth]{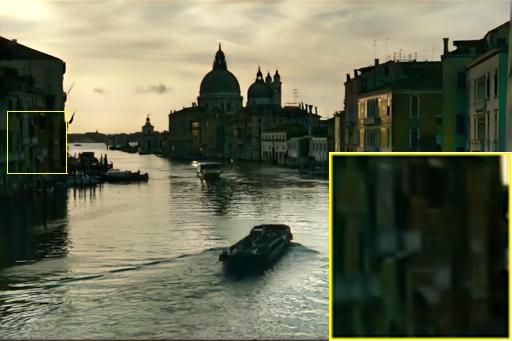}&\hspace{-4.5mm}
\includegraphics[width = 0.243\linewidth]{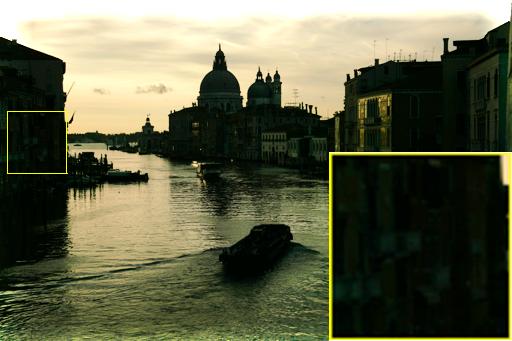}
\\
 (e) RRDNet~\cite{RRRDNet_zhu_icme2020}&\hspace{-4mm} (f) DeepUPE~\cite{deepupe_cvpr19}&\hspace{-4mm} (g) DRBN~\cite{DRBN_yang_cvpr20} &\hspace{-4mm}(h) FIDE~\cite{fide}
\\
\includegraphics[width = 0.243\linewidth]{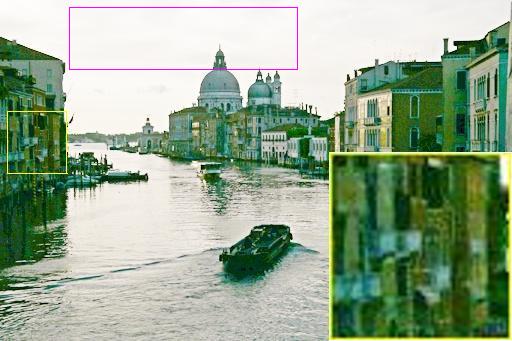}&\hspace{-4.5mm}
\includegraphics[width = 0.24\linewidth]{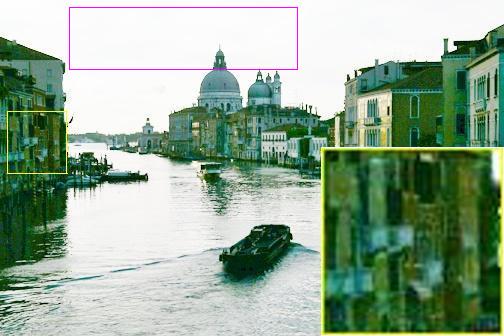}&\hspace{-4.5mm}
\includegraphics[width = 0.243\linewidth]{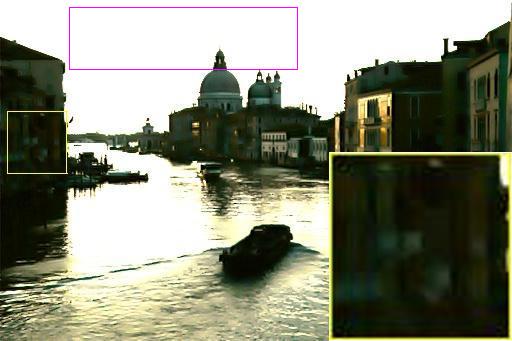}&\hspace{-4.5mm}
\includegraphics[width = 0.243\linewidth]{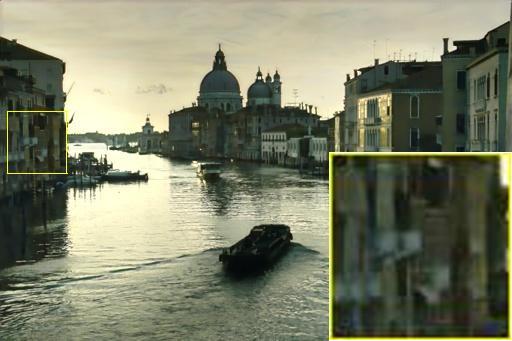}
\\
(i) Zero-DCE~\cite{zerodce_lowlight_guo} &\hspace{-4mm} (j) Zero-DCE++~\cite{zerodce_lowlight_pami}&\hspace{-4mm} (k) RUAS~\cite{RUAS_liu_cvpr21}&\hspace{-4mm} (l) SPGAT
\\
\end{tabular}
\end{center}
\caption{Comparisons with state-of-the-art methods on real-world images in the MEF dataset~\cite{mef}.
The proposed SPGAT is able to generate a clearer result with better details and texture.
}
\label{fig: Comparisons on MEF images.}
\end{figure*}

\begin{figure*}[!t]
\begin{center}
\begin{tabular}{ccccc}

\includegraphics[width = 0.243\linewidth]{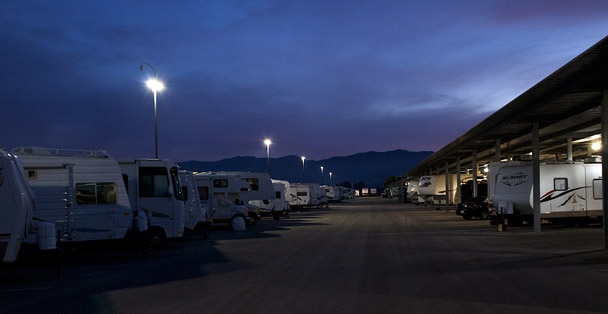}&\hspace{-4.5mm}
\includegraphics[width = 0.243\linewidth]{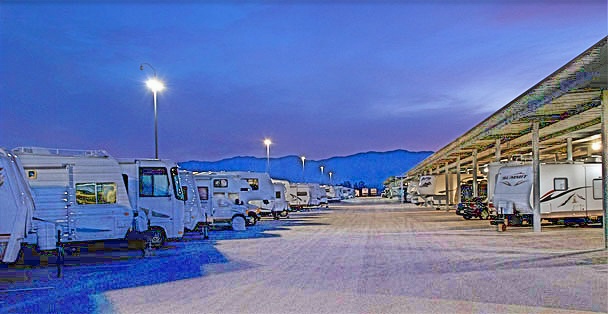}&\hspace{-4.5mm}
\includegraphics[width = 0.243\linewidth]{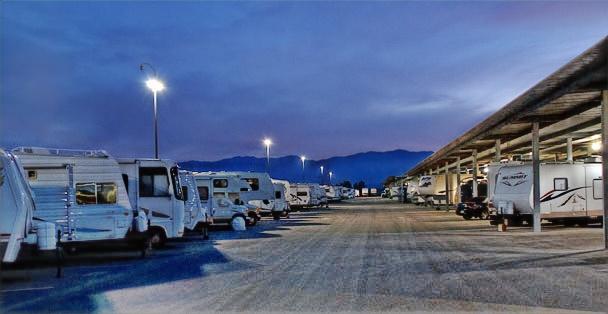}&\hspace{-4.5mm}
\includegraphics[width = 0.243\linewidth]{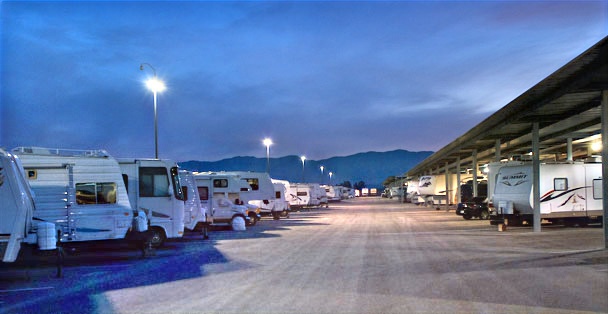}
\\
(a) Low-light &\hspace{-4mm} (b) Retinex~\cite{retinexnet_wei_bmvc18} &\hspace{-4mm} (c) KinD~\cite{kind_zhang_mm19} &\hspace{-4mm} (d) Enlighten~\cite{enlighten_tip_jiang}
\\
\includegraphics[width = 0.243\linewidth]{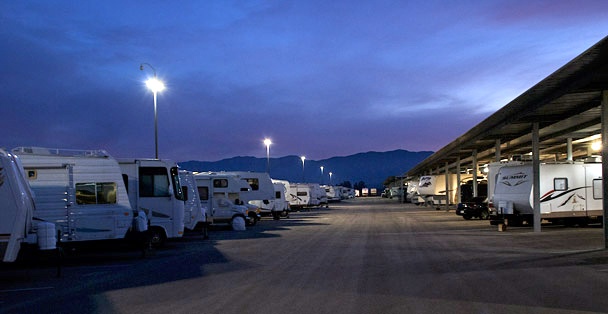}&\hspace{-4.5mm}
\includegraphics[width = 0.243\linewidth]{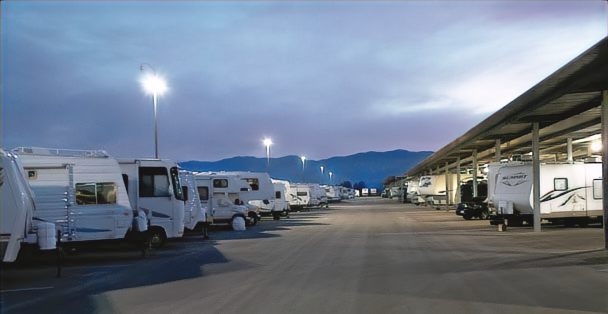}&\hspace{-4.5mm}
\includegraphics[width = 0.243\linewidth]{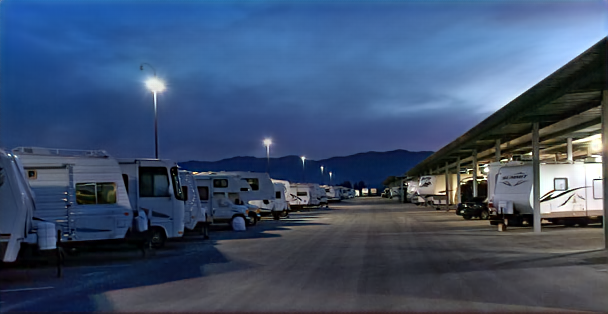}&\hspace{-4.5mm}
\includegraphics[width = 0.243\linewidth]{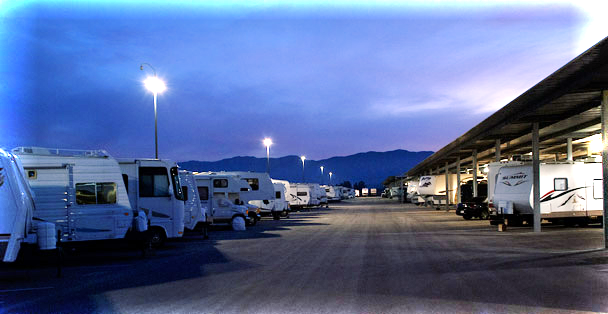}
\\
(e) RRDNet~\cite{RRRDNet_zhu_icme2020}&\hspace{-4mm} (f) DeepUPE~\cite{deepupe_cvpr19}&\hspace{-4mm} (g) DRBN~\cite{DRBN_yang_cvpr20} &\hspace{-4mm}(h) FIDE~\cite{fide}
\\
\includegraphics[width = 0.243\linewidth]{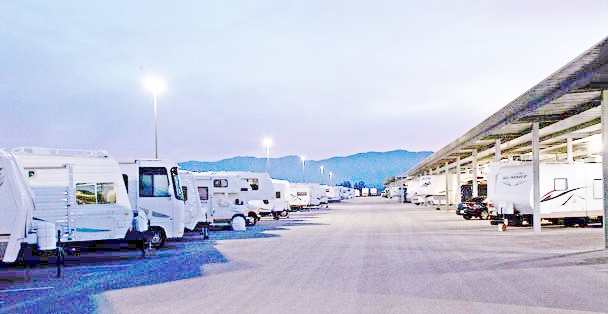}&\hspace{-4.5mm}
\includegraphics[width = 0.243\linewidth]{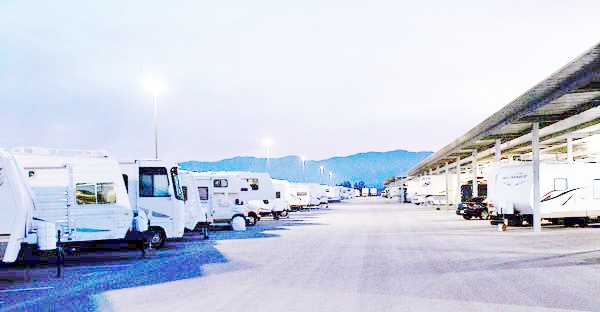}&\hspace{-4.5mm}
\includegraphics[width = 0.243\linewidth]{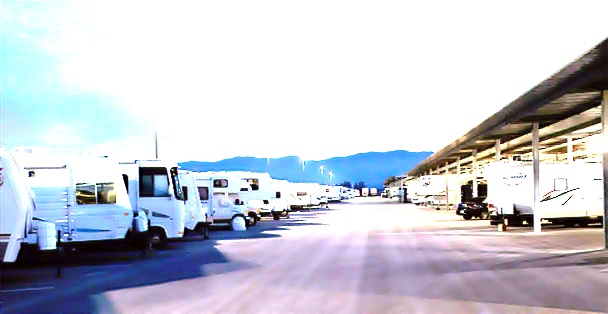}&\hspace{-4.5mm}
\includegraphics[width = 0.243\linewidth]{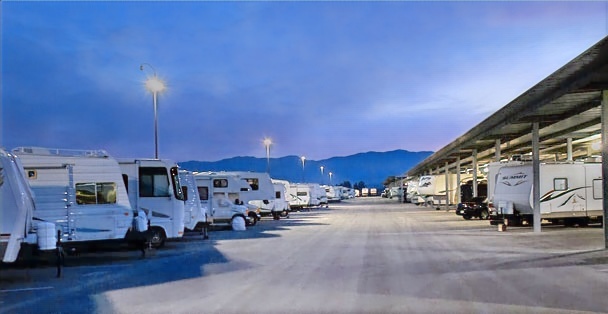}
\\
(i) Zero-DCE~\cite{zerodce_lowlight_guo} &\hspace{-4mm} (j) Zero-DCE++~\cite{zerodce_lowlight_pami}&\hspace{-4mm} (k) RUAS~\cite{RUAS_liu_cvpr21}&\hspace{-4mm} (l) SPGAT
\\
\end{tabular}
\end{center}
\caption{Comparisons with state-of-the-art methods on real-world images in the NPE dataset~\cite{NPE}.
The proposed SPGAT is able to generate a more realistic result.
}
\label{fig: Comparisons on NPE images.}
\end{figure*}

\begin{figure*}[!t]
\begin{center}
\begin{tabular}{ccccc}
\includegraphics[width = 0.243\linewidth]{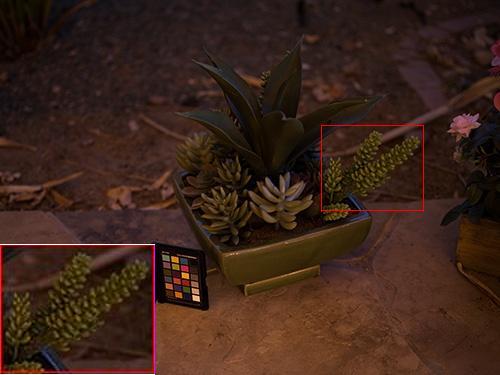}&\hspace{-4.5mm}
\includegraphics[width = 0.243\linewidth]{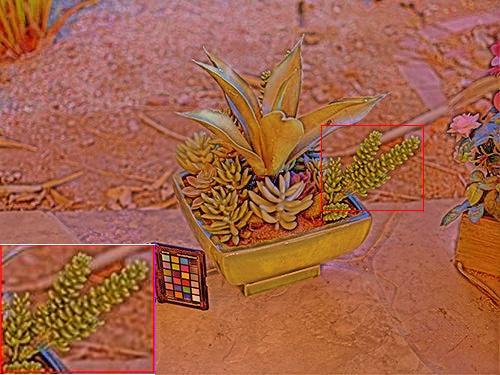}&\hspace{-4.5mm}
\includegraphics[width = 0.243\linewidth]{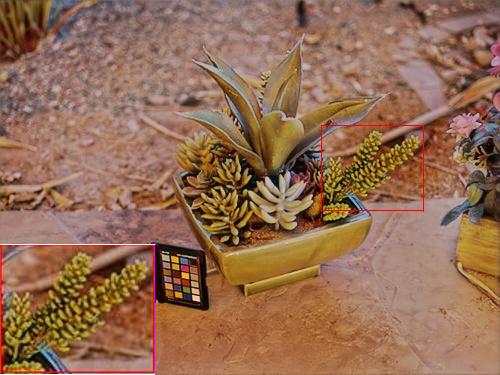}&\hspace{-4.5mm}
\includegraphics[width = 0.243\linewidth]{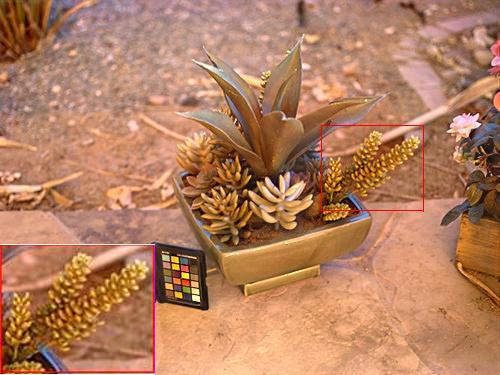}
\\
(a) Low-light &\hspace{-4mm} (b) Retinex~\cite{retinexnet_wei_bmvc18} &\hspace{-4mm} (c) KinD~\cite{kind_zhang_mm19} &\hspace{-4mm} (d) Enlighten~\cite{enlighten_tip_jiang}
\\
\includegraphics[width = 0.243\linewidth]{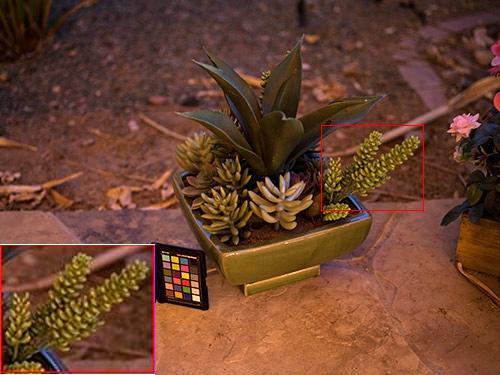}&\hspace{-4.5mm}
\includegraphics[width = 0.243\linewidth]{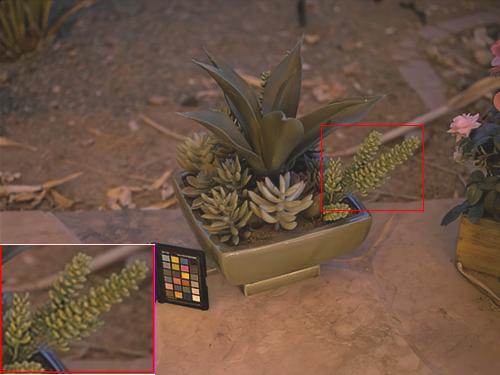}&\hspace{-4.5mm}
\includegraphics[width = 0.243\linewidth]{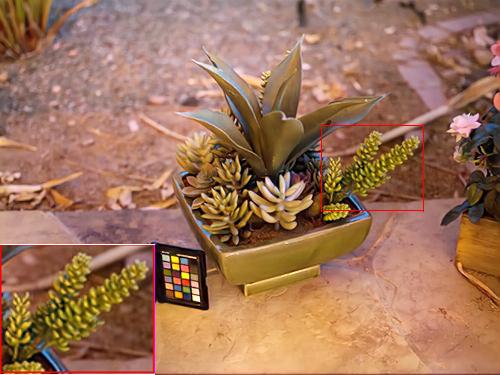}&\hspace{-4.5mm}
\includegraphics[width = 0.243\linewidth]{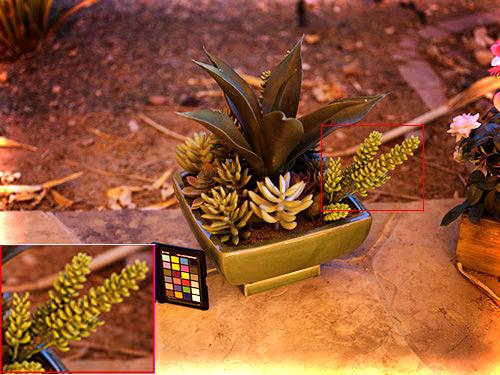}
\\
 (e) RRDNet~\cite{RRRDNet_zhu_icme2020}&\hspace{-4mm} (f) DeepUPE~\cite{deepupe_cvpr19}&\hspace{-4mm} (g) DRBN~\cite{DRBN_yang_cvpr20} &\hspace{-4mm}(h) FIDE~\cite{fide}
\\
\includegraphics[width = 0.243\linewidth]{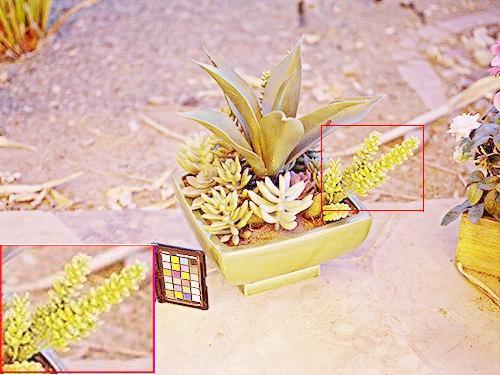}&\hspace{-4.5mm}
\includegraphics[width = 0.243\linewidth]{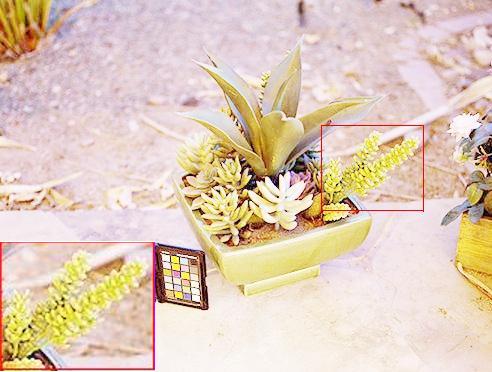}&\hspace{-4.5mm}
\includegraphics[width = 0.243\linewidth]{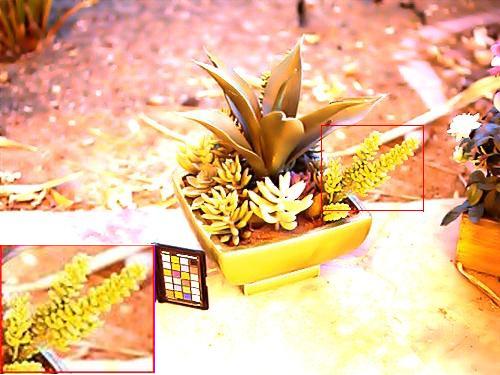}&\hspace{-4.5mm}
\includegraphics[width = 0.243\linewidth]{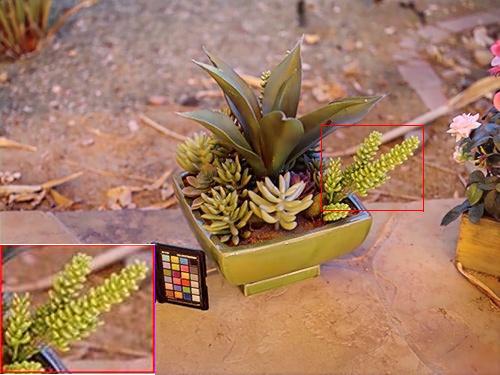}
\\
(i) Zero-DCE~\cite{zerodce_lowlight_guo} &\hspace{-4mm} (j) Zero-DCE++~\cite{zerodce_lowlight_pami}&\hspace{-4mm} (k) RUAS~\cite{RUAS_liu_cvpr21}&\hspace{-4mm} (l) SPGAT
\\

\end{tabular}
\end{center}
\caption{Comparisons with state-of-the-art methods on real-world images in the LIME dataset~\cite{lime_Retinex_tip17}.
The proposed SPGAT is able to generate a more natural result, especially in the zooming-in region.
}
\label{fig: Comparisons on LIME images.}
\end{figure*}
\subsection{Datasets and Evaluation Criteria}
\subsubsection{Synthetic Datasets}
LOL dataset~\cite{retinexnet_wei_bmvc18} is a widely used dataset, which contains 485 training samples and 15 testing samples.
Wei~\etal. in \cite{retinexnet_wei_bmvc18} also collect 1000 raw images from RAISE~\cite{raise} as normal-light images $\mathrm{E}$ and use them to synthesize low-light images $\mathrm{L}$.
We name this dataset Brightening.
In this dataset, 900 images are used for training, and the remaining 100 images are used for testing.
We use the two datasets to evaluate the enhancement performance of synthetic images.
Moreover, we use gradient operation on $\mathrm{L}$ to obtain the input structure $\mathrm{S}$ and on $\mathrm{E}$ to produce the ground truth of the structure $\mathrm{P}$.
\subsubsection{Real-World Datasets}
DICM~\cite{dicm}, LIME~\cite{lime_Retinex_tip17}, MEF~\cite{mef}, and NPE~\cite{NPE} are widely used real-world datasets.
We use them to evaluate the enhanced quality of real-world scenarios.
\subsubsection{Evaluation Criteria}
Peak Signal to Noise Ratio (PSNR)~\cite{PSNR_thu} and Structural Similarity Index Measure (SSIM)~\cite{SSIM_wang} are two widely used metrics to measure the enhanced results with Ground-Truth (GT) object.
We use them to evaluate the quality of restored images on synthetic datasets.
As there are no ground-truth normal-light images for real-world low-light ones, we only compare the results visually.

\subsection{Results on Synthetic Datasets}
We first evaluate our method against state-of-the-art ones on synthetic datasets. For fair comparisons, we retrain the deep learning-based methods using the same training datasets as the proposed method.

Tab.~\ref{tab:syn-Results-sota} summarises the quantitative results in the LOL dataset, where our approach outperforms state-of-the-art methods in terms of PSNR and SSIM.
Fig.~\ref{fig: Comparisons on the LOL dataset3.} presents one example from the LOL dataset.
Our method is able to generate a more natural result with better texture in the zooming-in region.
Tab.~\ref{tab:Comparisons with baselines on the Brightening dataset.} reports the enhancement results in the Brightening dataset.
Our approach also achieves the best performance in the dataset.
Fig.~\ref{fig: Comparisons on the bri dataset2.} provides two examples from the Brightening dataset.
The Zero-DCE~\cite{zerodce_lowlight_guo} always generates the results with color distortions.
Our approach produces a globally brighter result with better textures in the cropped region, while other state-of-the-art methods produce locally under-enhancement results.

%
\subsection{Results on Real-World Images}
We then evaluate our method on the real-world images in  Fig.~\ref{fig: Comparisons on DICM images.}, Fig.~\ref{fig: Comparisons on MEF images.}, Fig.~\ref{fig: Comparisons on NPE images.}, and Fig.~\ref{fig: Comparisons on LIME images.}.
For fair comparisons, all the real-world image enhancement results are produced by the models trained on the LOL dataset.
Fig.~\ref{fig: Comparisons on DICM images.} and Fig.~\ref{fig: Comparisons on MEF images.} illustrate that our method is able to produce clearer results with finer texture.
Note that Zero-DCE~\cite{zerodce_lowlight_guo}, Zero-DCE++~\cite{zerodce_lowlight_pami}, and RUAS~\cite{RUAS_liu_cvpr21} always generate over-enhancement results.
Results in Fig.~\ref{fig: Comparisons on NPE images.} reveal that our approach can generate a cleaner result, while other state-of-the-art methods, e.g., DRBN, produce under-enhancement quality.
Fig.~\ref{fig: Comparisons on LIME images.} shows that our proposed SPGAT produces a more natural result, especially in the zooming-in region.
%
These examples in diverse real-world datasets have adequately demonstrated that our model generates much clearer images which look more natural, demonstrating the effectiveness and better generalization ability of the proposed method in real-world conditions.

\begin{table*}[!t]
\centering
\caption{Ablation study on basic component.
The \CheckmarkBold and \XSolidBrush denote that the corresponding component is respectively adopted and not adopted.
The results reveal that the concatenation operation is a better manner of skip connection between encoder and decoder in Transformers, while both the proposed structural prior guidance \eqref{eq:spgm} and adversarial learning  \eqref{eq:lossencoder}\&\eqref{eq:lossdecoder} can help improve the enhancement quality.
}
\scalebox{1}{
\centering
\begin{tabular}{lccccccc}
\toprule
Experiments & $M_{1}$  & $M_{2}$ & $M_{3}$  &  $M_{4}$ &  $M_{5}$      & $M_{6}$  & $M_{7}$ (Ours)    \\
\toprule
Without skip connection               & \CheckmarkBold   & \XSolidBrush & \XSolidBrush   & \XSolidBrush & \XSolidBrush & \XSolidBrush  & \XSolidBrush     \\
Skip connection by summation             &\XSolidBrush   & \CheckmarkBold & \CheckmarkBold & \XSolidBrush & \XSolidBrush    & \XSolidBrush  &   \XSolidBrush\\
Skip connection by concatenation   & \XSolidBrush   & \XSolidBrush & \XSolidBrush& \CheckmarkBold & \CheckmarkBold    & \CheckmarkBold  &   \CheckmarkBold\\
Structural prior guidance by \eqref{eq:concat}  & \XSolidBrush   & \XSolidBrush& \XSolidBrush & \XSolidBrush &\CheckmarkBold & \XSolidBrush  &   \XSolidBrush   \\
Structural prior guidance by  \eqref{eq:spgm}     &\XSolidBrush   & \XSolidBrush& \CheckmarkBold & \XSolidBrush&\XSolidBrush & \CheckmarkBold   &   \CheckmarkBold   \\
Adversarial learning   \eqref{eq:lossencoder}\&\eqref{eq:lossdecoder}            &\XSolidBrush   & \XSolidBrush& \XSolidBrush   & \XSolidBrush&\XSolidBrush & \XSolidBrush  &   \CheckmarkBold   \\
\bottomrule
PSNR $\uparrow$                 & 13.08          & 19.26&19.60            &19.40 & 18.10             & 19.68           & \textbf{19.80}           \\
SSIM $\uparrow$                   &0.5054       &0.8134 &0.8154           & 0.8160& 0.7920             & 0.8187         & \textbf{0.8234}     \\
\bottomrule
\end{tabular}}
\label{tab:Ablation study on basic component of the model.}
\end{table*}

\begin{figure*}[!t]
\begin{center}
\begin{tabular}{ccccc}
\includegraphics[width = 0.195\linewidth]{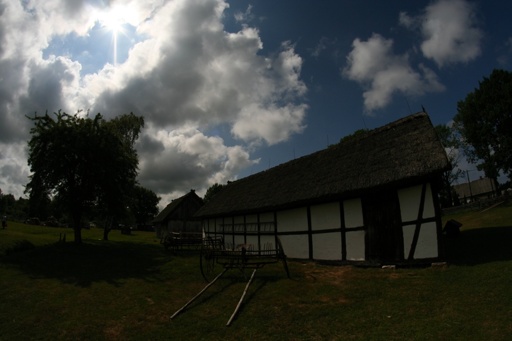}&\hspace{-4.5mm}
\includegraphics[width = 0.195\linewidth]{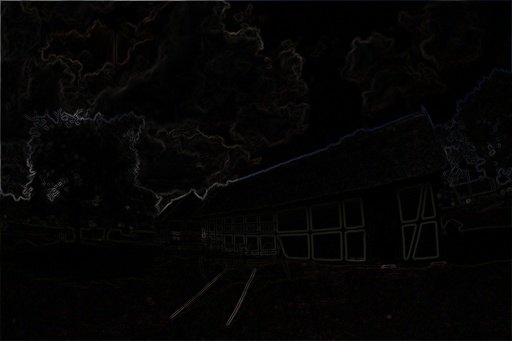}&\hspace{-4.5mm}
\includegraphics[width = 0.195\linewidth]{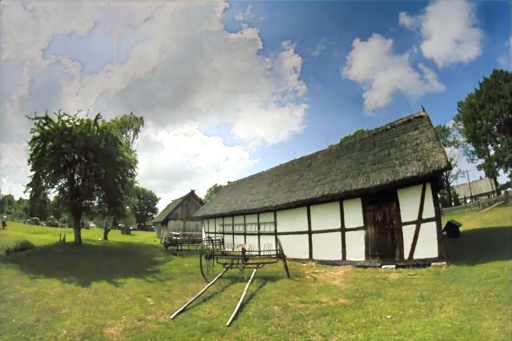}&\hspace{-4.5mm}
\includegraphics[width = 0.195\linewidth]{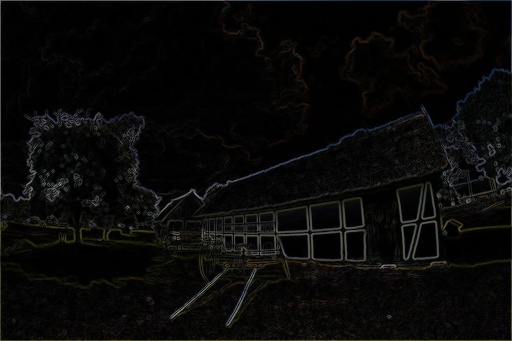}&\hspace{-4.5mm}
\includegraphics[width = 0.195\linewidth]{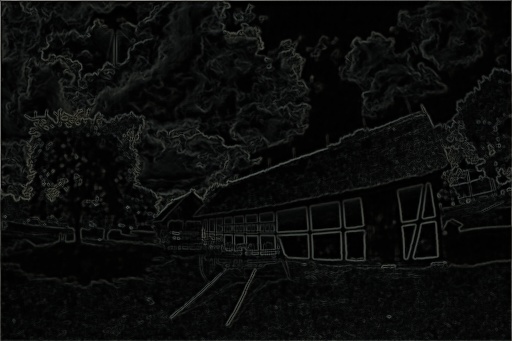}
\\
(a)  &\hspace{-4.5mm} (b)  &\hspace{-4.5mm} (c)  &\hspace{-4.5mm} (d) &\hspace{-4.5mm} (e)

\end{tabular}
\end{center}
\caption{Accurate estimation of SPE on a real-world example.
(a) low-light;
(b) structure of (a);
(c) output of generator;
(d) structure of (c);
(e) output of SPE.
The structural prior estimator is able to accurately estimate image structure (e).
}
\label{fig: A real-world enhanced example and corresponding structure result.}
\end{figure*}

\begin{figure*}[!t]
\begin{center}
\begin{tabular}{ccccc}
\includegraphics[width = 0.195\linewidth]{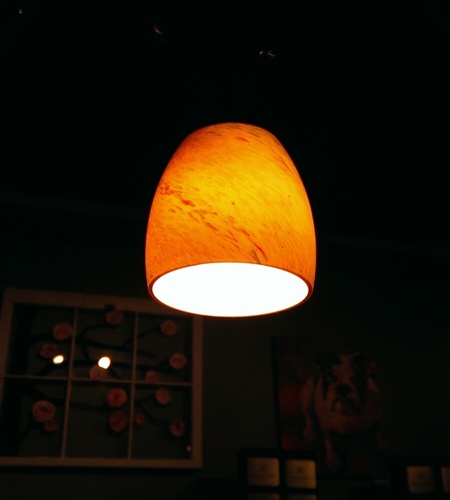}&\hspace{-4.5mm}
\includegraphics[width = 0.195\linewidth]{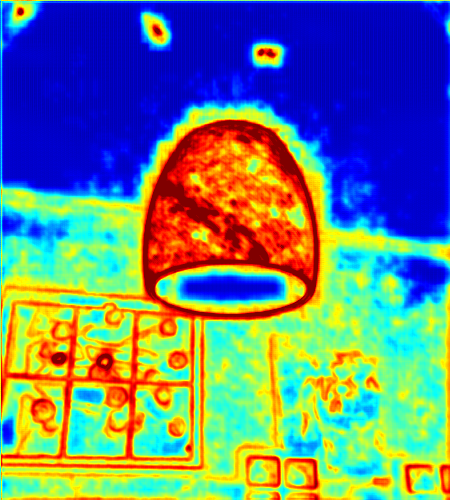}&\hspace{-4.5mm}
\includegraphics[width = 0.195\linewidth]{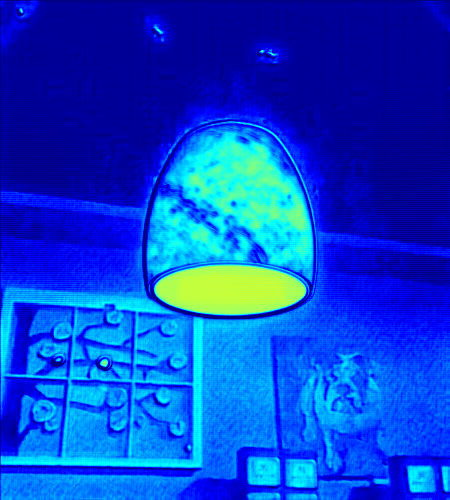}&\hspace{-4.5mm}
\includegraphics[width = 0.195\linewidth]{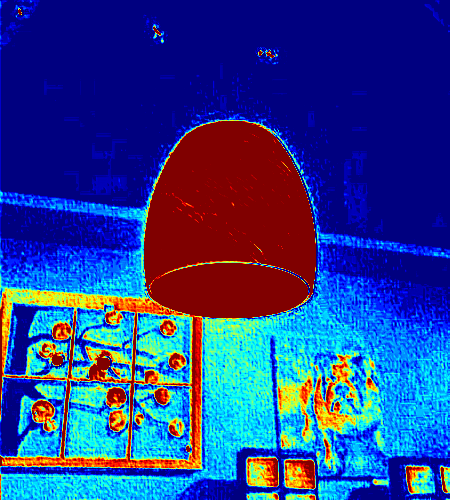}&\hspace{-4mm}
\includegraphics[width = 0.195\linewidth]{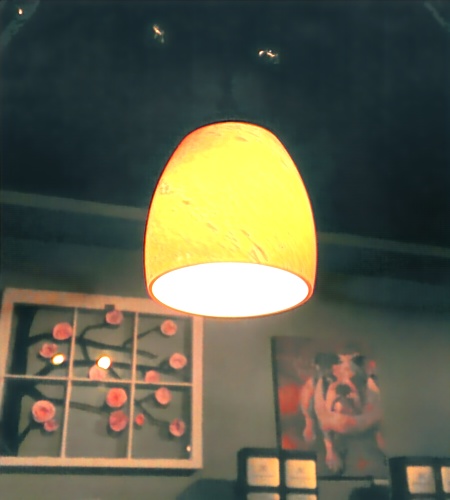}
\\
(a)  &\hspace{-4.5mm} (b)  &\hspace{-4.5mm} (c)  &\hspace{-4.5mm} (d)&\hspace{-4.5mm} (e)

\end{tabular}
\end{center}
\caption{Effectiveness of the SPE on a real-world example.
(a) low-light input;
(b) feature from SPE;
(c) feature before SPGM in the generator;
(d) feature after SPGM in the generator;
(d) enhanced image.
The SPE can help the generator pay more attention to the structural content (d).
}
\label{fig: A visualization example on the effect for SPB.}
\end{figure*}

%
\subsection{Analysis and Discussions}
In this section, we demonstrate the effectiveness of each component of the proposed method. All the baselines in this section are trained using the same settings as the proposed method for fair comparisons.
\subsubsection{Analysis on the Basic Components}\label{sec: Analysis on Basic Component}
We first demonstrate the effectiveness of skip connection (that is used in both generator and SPE), structural prior, and adversarial learning.
\begin{figure}[!t]
\begin{center}
\begin{tabular}{ccccc}
\includegraphics[width = 0.322\linewidth]{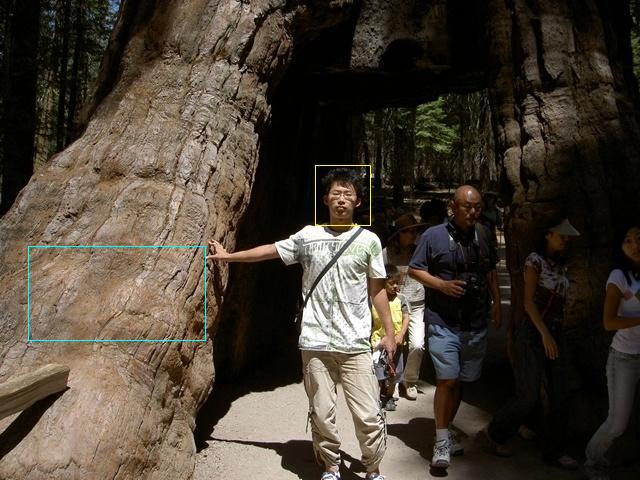}&\hspace{-4.5mm}
\includegraphics[width = 0.322\linewidth]{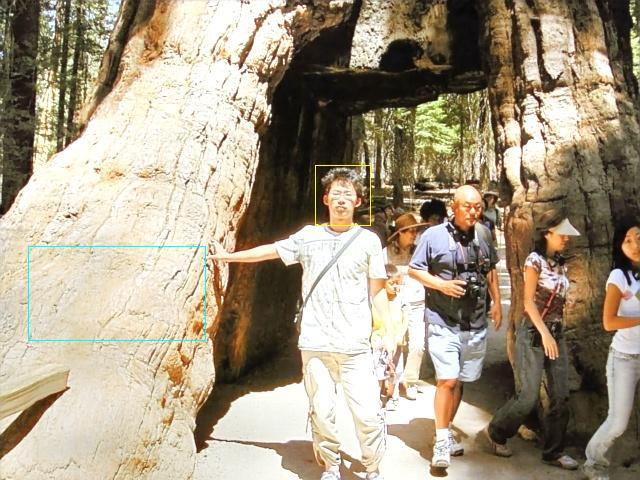}&\hspace{-4.5mm}
\includegraphics[width = 0.322\linewidth]{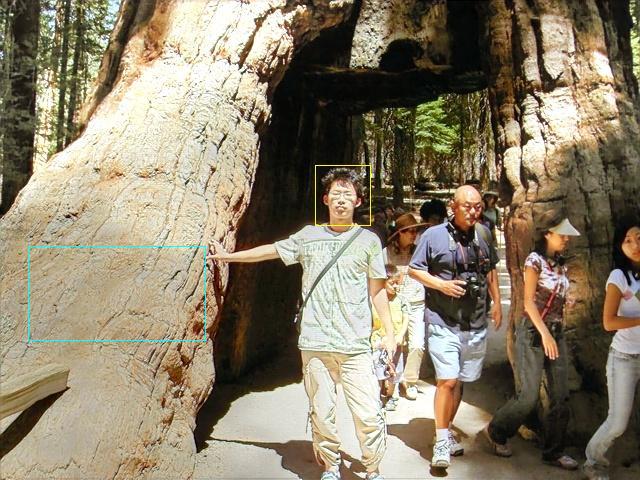}
\\
(a)  &\hspace{-4mm}  (b)&\hspace{-4mm} (c)
\\
\end{tabular}
\end{center}
\caption{Effectiveness of SPE on a real-world example.
(a) low-light;
(b) the result of the model without SPE;
(c) the result of the model with SPE.
The SPE can refrain the generator from over-enhancement results so that it can help the network produce a better result with finer structure (c).
}
\label{fig: A real-world enhanced example on the effect of SPB.}
\end{figure}

\begin{figure}[!t]
\begin{center}
\begin{tabular}{ccc}
\includegraphics[width = 0.322\linewidth]{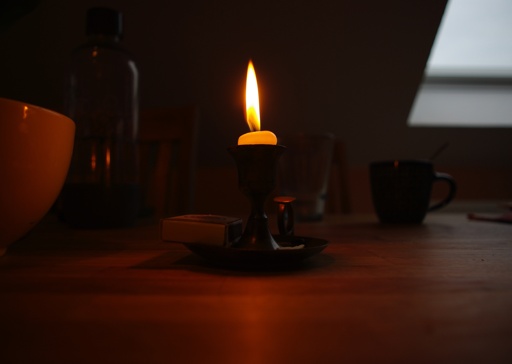}&\hspace{-4.5mm}
\includegraphics[width = 0.322\linewidth]{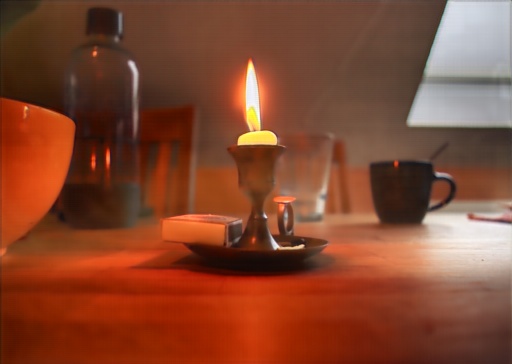}&\hspace{-4.5mm}
\includegraphics[width = 0.322\linewidth]{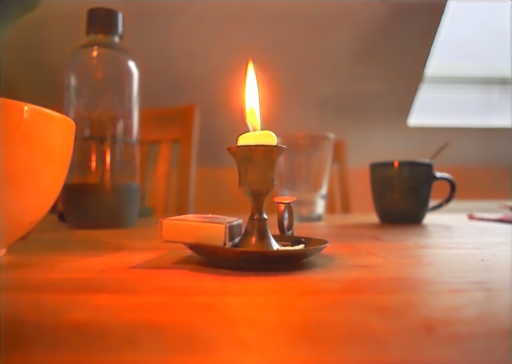}
\\
(a)  &\hspace{-4mm}  (b)&\hspace{-4mm} (c)

\end{tabular}
\end{center}
\caption{Effect of adversarial learning on one real-world image.
(a) low-light input;
(b) the result of the model without adversarial learning;
(c) the result of the model with adversarial learning.
Adversarial learning can help generate a more natural result (c).
}
\label{fig: Comparisons on whether with adversarial learning.}
\end{figure}

Tab.~\ref{tab:Ablation study on basic component of the model.} shows that the skip connection can significantly improve the results ($M_{1}$ vs. $M_{2}$ and $M_{4}$), while concatenation operation generates better results than the element-wise summation ($M_{2}$ vs. $M_{4}$).
Furthermore, we verify that structural prior can help improve the enhancement performance ($M_{2}$ vs. $M_{3}$ and $M_{4}$ vs. $M_{6}$).
Moreover, another alternative guided model to replace \eqref{eq:spgm} is that we cascade the features in SPE and the features in generator, which can be expressed as:
\begin{equation}
\begin{array}{ll}
\mathrm{SPGM}_{\mathrm{Concat}}(\mathbf{F}_{\mathrm{E}}) = \mathrm{Linear}\big(\mathrm{Concat}[\mathbf{F}_{\mathrm{P}}, \mathbf{F}_{\mathrm{E}}]\big),
\end{array}
\label{eq:concat}
\end{equation}
where $\mathrm{Linear}$ denotes the linear layer that is to convert the concatenation dimension to the original dimension, while $\mathrm{Concat}$ refers to the concatenation operation.
We also find that the proposed guided way \eqref{eq:spgm} without extra parameters outperforms against the concatenation \eqref{eq:concat} with consuming parameters ($M_{5}$ vs. $M_{6}$, 1.58 dB gain).
\begin{table*}[!t]
\centering
\caption{Ablation study on the discriminators. $\mathrm{G}$ and $\mathrm{D}$ denotes the generator and discriminators, respectively.
It is notable that the results reveal that the model with a traditional single discriminator (the first column) is worse than the model without adversarial learning ($M_{6}$ in Tab.~\ref{tab:Ablation study on basic component of the model.}), while our final model (the last column) with skip connection between the generator and discriminators and the guidance of structural prior outperforms both.
}
\scalebox{1.0}{
\begin{tabular}{lccccccc}
\toprule
Skip connection between $\mathrm{G}$ and $\mathrm{D}$& \XSolidBrush   & \XSolidBrush &\CheckmarkBold  &\CheckmarkBold& \CheckmarkBold &  \CheckmarkBold     \\
Structural prior guidance& \XSolidBrush   & \XSolidBrush     & \XSolidBrush    & \CheckmarkBold &   \XSolidBrush &   \CheckmarkBold         \\
Single discriminator&\CheckmarkBold  & \XSolidBrush & \CheckmarkBold    & \CheckmarkBold&   \XSolidBrush  &   \XSolidBrush  \\
Dual discriminators  &\XSolidBrush   & \CheckmarkBold     &\XSolidBrush & \XSolidBrush            &\CheckmarkBold &   \CheckmarkBold   \\
\bottomrule
PSNR $\uparrow$& 19.40&19.51  & 19.45       & 19.49     &19.57 & \textbf{19.80}           \\
SSIM $\uparrow$&0.8186 &0.8193   & 0.8188       & 0.8190  &0.8198  & \textbf{0.8234}     \\
\bottomrule
\end{tabular}}
\label{tab:Ablation study on discriminator.}
\end{table*}
\begin{figure*}[!t]
\begin{center}
\begin{tabular}{ccccccc}
\includegraphics[width = 0.24\linewidth]{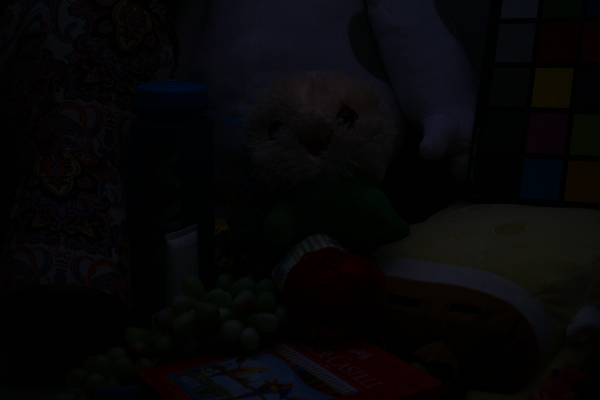}&\hspace{-4.5mm}
\includegraphics[width = 0.24\linewidth]{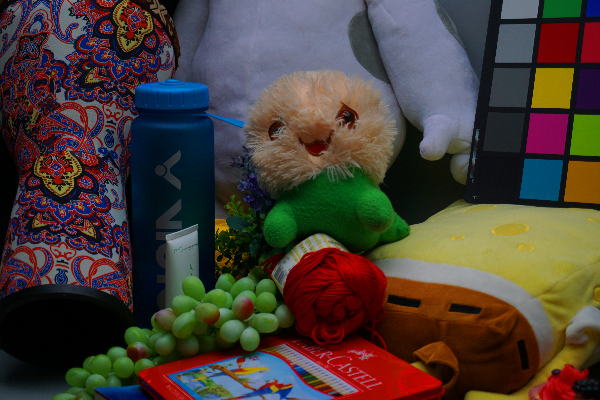}&\hspace{-4.5mm}
\includegraphics[width = 0.24\linewidth]{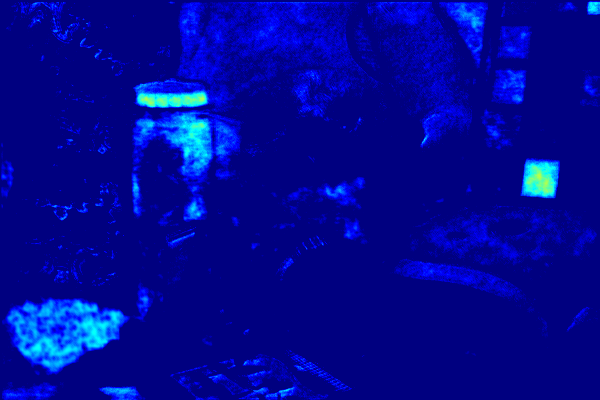}&\hspace{-4.5mm}
\includegraphics[width = 0.24\linewidth]{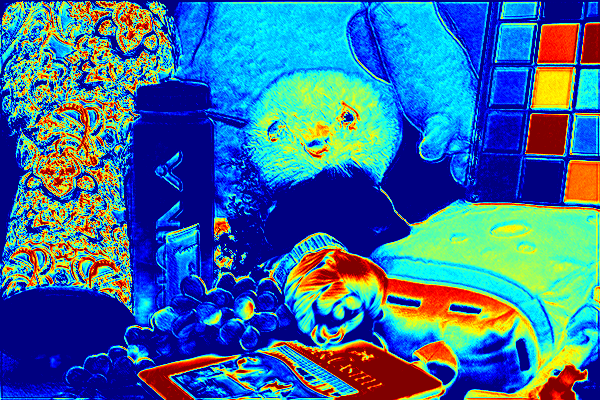}
\\
(a)  &\hspace{-4.5mm} (b)  &\hspace{-4.5mm} (c)  &\hspace{-4.5mm} (d)
\\

\includegraphics[width=3.3cm, height=3cm]{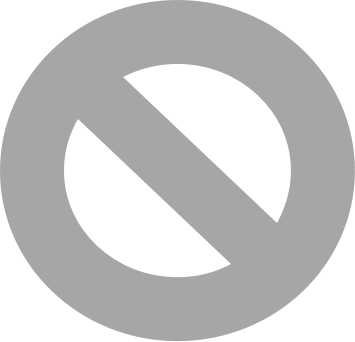}&\hspace{-4.5mm}
\includegraphics[width = 0.24\linewidth]{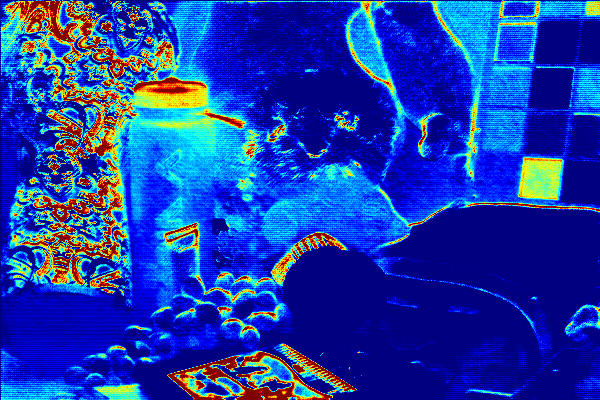}&\hspace{-4.5mm}
\includegraphics[width = 0.24\linewidth]{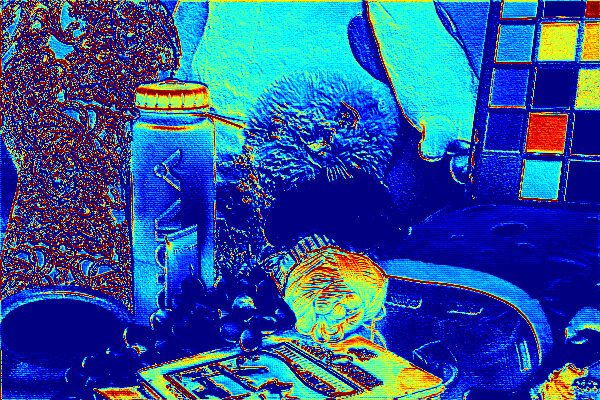} &\hspace{-4.5mm}
\includegraphics[width = 0.24\linewidth]{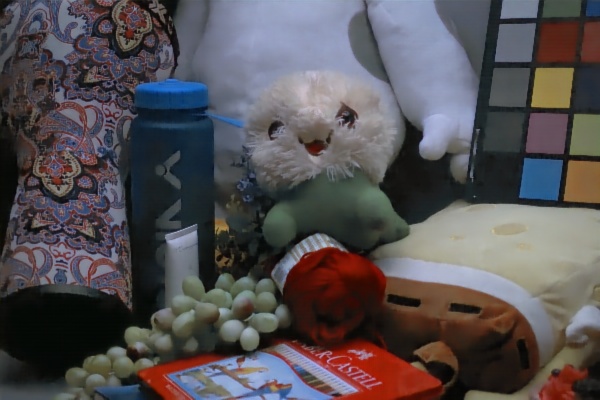}
\\

 &\hspace{-4.5mm}(e) &\hspace{-4.5mm} (f) &\hspace{-4.5mm} (g)
\\
\end{tabular}
\vspace{-2mm}
\caption{Feature visualization at the encoder and decoder stage on the low-light input and corresponding normal-light input.
(a) low-light input;
(b) normal-light input;
(c) reconstructed encoder feature;
(d) normal-light encoder feature;
(e) reconstructed decoder feature;
(f) normal-light decoder feature;
(g) our result.
The reconstructed feature (c) at the encoder stage is vague, while the reconstructed feature (e) at the decoder stage is able to produce a clearer outline.
The reconstructed feature (c) is quite different from that of the normal-light feature (d) in the encoder stage, while the difference ((e) and (f)) becomes smaller at the decoder stage.
Hence, utilizing two discriminators to respectively discriminate encoder and decoder features can better measure the difference between reconstructed features and normal-light ones for better image restoration.
}
\label{fig: Feature visualization at encoder and decoder stage for the effect on low-light and corresponding normal-light input.}
\end{center}
\end{figure*}
\begin{figure*}[!t]
\begin{center}
\begin{tabular}{ccccc}
\includegraphics[width = 0.24\linewidth]{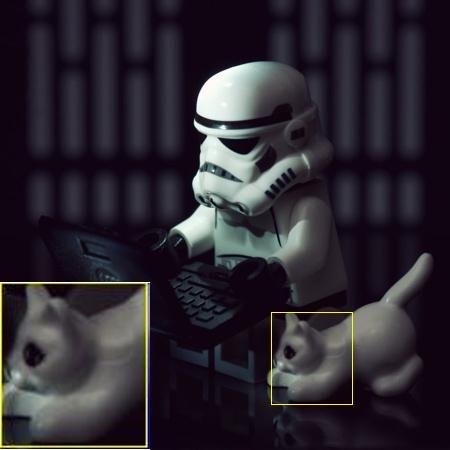}&\hspace{-4.5mm}
\includegraphics[width = 0.24\linewidth]{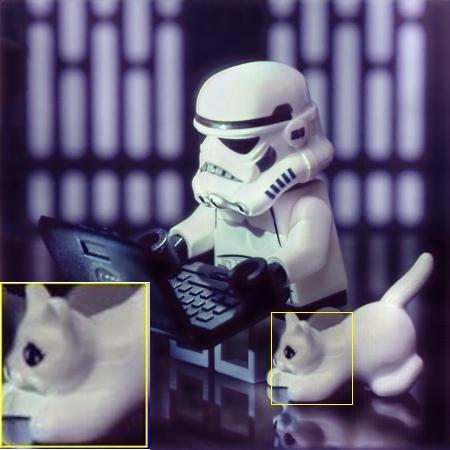}&\hspace{-4.5mm}

\includegraphics[width = 0.24\linewidth]{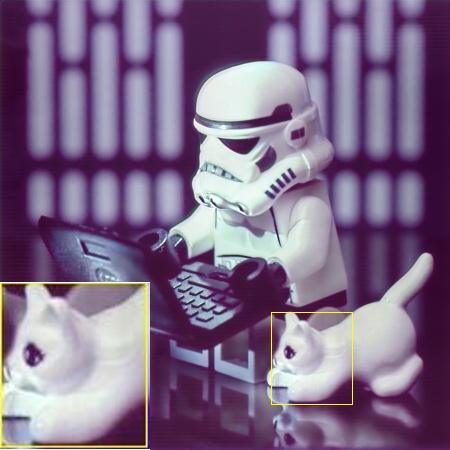}&\hspace{-4.5mm}
\includegraphics[width = 0.24\linewidth]{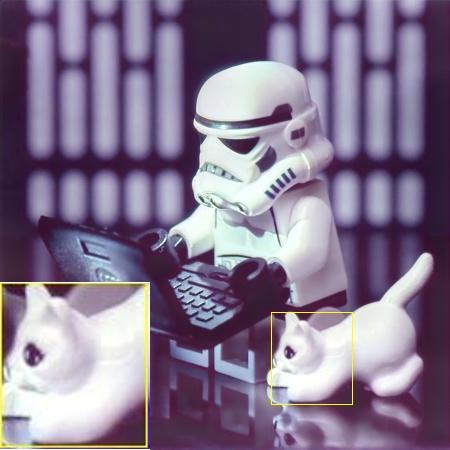}
\\

 (a) &\hspace{-4mm}  (b)  &\hspace{-4mm} (c) &\hspace{-4mm} (d) \\
\end{tabular}
\end{center}
\vspace{-2mm}
\caption{Effect on the structural prior and skip connection between the generator and discriminators on one real-world image.
(a) low-light input;
(b) result of the model without skip connection between the generator and discriminators;
(c) result of the model with skip connection between the generator and discriminators;
(d) result of the model with skip connection between the generator and discriminators, and with structural prior guidance.
Both the skip connections between the generator and discriminators and the guidance of structural prior can improve image enhancement quality.
}
\label{fig: Effect on the skip connection between generator and discriminators.}
\end{figure*}
As shown in Fig.~\ref{fig: A real-world enhanced example and corresponding structure result.}, SPE is capable of accurately generating image structure (Fig.~\ref{fig: A real-world enhanced example and corresponding structure result.}(e)), which thus can provide the generator and discriminators better priors for better image enhancement.
Fig.~\ref{fig: A visualization example on the effect for SPB.} presents a visualization example of the effect of SPE.
%
The SPE is able to generate a more distinct structure (Fig.~\ref{fig: A visualization example on the effect for SPB.}(b)) so which provides a positive effect to help the generator pay more attention to structure after SPGM (Fig.~\ref{fig: A visualization example on the effect for SPB.}(d)).
%
Fig.~\ref{fig: A real-world enhanced example on the effect of SPB.} presents a visual example on the effect of SPE.
We can observe that the model without SPE tends to lose some details, while SPE is able to help preserve the better structural details.

Furthermore, we also observe in Tab.~\ref{tab:Ablation study on basic component of the model.} that our proposed adversarial learning manner is able to further improve enhancement results ($M_{6}$ vs. $M_{7}$).
We also present a real-world visual example of the effect of adversarial learning in Fig.~\ref{fig: Comparisons on whether with adversarial learning.}, which shows that adversarial learning helps generate a more natural result.
These experiments demonstrate that the designed components are beneficial to image enhancement.

\subsubsection{Analysis on the Discriminators}\label{sec: Analysis on the Discriminator}

One may wonder why we design two discriminators.
To answer this question, we visualize the features\footnote{We ensure that the features are generated at the same position at the encoder or decoder stage.} at the encoder and decoder stage for the effect on low-light and corresponding normal-light input in Fig.~\ref{fig: Feature visualization at encoder and decoder stage for the effect on low-light and corresponding normal-light input.}.
We observe that the constructed feature at the encoder stage is vague, while the reconstructed feature at the decoder stage is able to produce a clearer outline.
This is the reason why we use two discriminators as the features between the encoder and decoder stages are much different.
Moreover, we also observe that the reconstructed feature is quite different from that of the normal-light features in the encoder stage, while the difference becomes smaller at the decoder stage.
Hence, utilizing two discriminators to respectively discriminate encoder and decoder features can better measure the difference between reconstructed features and normal-light ones for better image restoration.

Furthermore, as the two discriminators employ the skip connections between the generator and discriminators with the guidance by structural prior to guide the discriminating process, we need to examine the effect of these operations.
Tab.~\ref{tab:Ablation study on discriminator.} reports the ablation results.
We can observe that the dual discriminators indeed produce better results than a single discriminator, while the skip connections between the generator and discriminators can further improve the performance.
Note that the structural prior that guides the discriminating process in the features from generator and discriminators improves the enhancement results.
These experiments demonstrate that the proposed structural prior guided discriminators with skip connections between the generator and discriminators are effective.

\begin{table}[!t]
\centering
\caption{Ablation study on the different manners of structure prior.
\textit{HPF} denotes the High-Pass Filtering operation.
}
	\scalebox{0.99999}{
		
		\begin{tabular}{lccccccc}
\toprule
Input image as structure prior    &\CheckmarkBold  & \XSolidBrush  &  \XSolidBrush\\
\textit{HPF}(Input image) as structure prior & \XSolidBrush& \CheckmarkBold&   \XSolidBrush\\
Gradient as structure prior& \XSolidBrush   & \XSolidBrush & \CheckmarkBold  \\
\bottomrule
PSNR $\uparrow$  & 15.95           &19.24   & \textbf{19.80}                     \\
SSIM $\uparrow$  &0.7135       &0.8134      & \textbf{0.8234}            \\
\bottomrule
\end{tabular}}
\label{tab:Ablation study on different manners of structure prior.}
\end{table}

Fig.~\ref{fig: Effect on the skip connection between generator and discriminators.} shows the effect of the skip connections between generator and discriminators.
%
We note that the proposed skip connection guided by structural prior in discriminators generates a better result as the skip connections between the generator and discriminators can provide the discriminators with more discriminative features so that the discriminators can better discriminate to help the generator better image restoration.
Moreover, the structural prior can further help the discriminators obtain structures from SPE for facilitating to produce better-enhanced images.

\subsubsection{Analysis on the Different Structure Priors}\label{sec: Analysis on Different Structure Priors}
One may want to know which structural prior is better for enhancement.
To answer this question, we use different manners to obtain the structural prior and the results are reported in Tab.~\ref{tab:Ablation study on different manners of structure prior.}.

We can observe that the input image as the structure prior cannot help generate satisfactory results, while we also note that the high-pass filtered image as the structural prior produces better performance than the model with the image as structure.
Meanwhile, we find that gradient as structure prior obtains the best enhancement results.
Hence, we use the image gradient to produce the structure.

\begin{table}[!t]
\centering
\caption{Ablation study on the combination of windows.
}
	\scalebox{0.99}{
\centering
\begin{tabular}{lccccccc}
\toprule
Combination of windows  &    PSNR $\uparrow$  & SSIM $\uparrow$
\\
\toprule
Single window: \{2\}& 19.55  &0.8158 \\
Single window: \{4\}&19.56    &0.8171 \\
Single window: \{8\}&19.59 & 0.8159 \\
Multiple parallel windows: \{2, 2, 2\}& 19.30    & 0.8175\\
Multiple parallel windows:  \{4, 4, 4\}& 19.60  & 0.8179 \\
Multiple parallel windows:  \{8, 8, 8\}&19.33 & 0.8185\\
\toprule
Multiple parallel windows:  \{2, 4, 8\} (Ours)& \textbf{19.80}& \textbf{0.8234}  \\

\toprule
\end{tabular}}
\label{tab:Ablation study on the combination of different windows.}
\end{table}

\subsubsection{Analysis on the Combination of Windows}\label{sec: Analysis on Combination of Different Windows}
As we use the parallel windows-based Swin Transformer block to replace the single window in ~\cite{Liu_swintrans_ICCV2021}, we are necessary to analyze its effect.
Tab.~\ref{tab:Ablation study on the combination of different windows.} reports the results.
The combination of multi-windows with the same window size is able to help produce higher SSIM results than the model with a single window.
Note that our proposed parallel windows with different window sizes achieve the best performance than other manners.
As each window can capture different content, fusing these different level hierarchical features in parallel windows can further improve the representation ability of the Transformer.

\begin{figure}[!t]
\begin{center}
\begin{tabular}{ccccc}

\includegraphics[width = 1\linewidth]{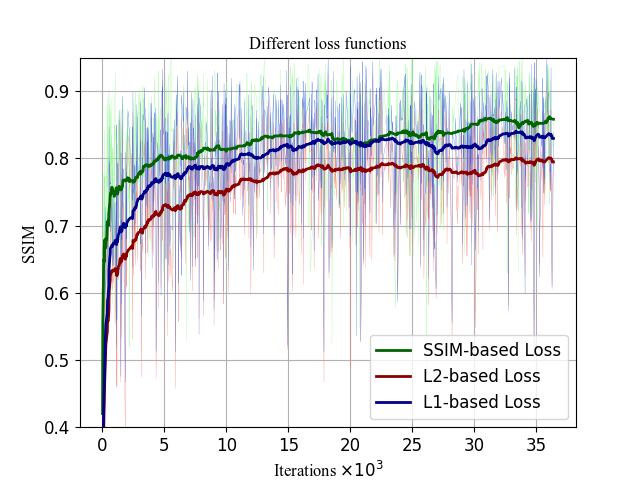}
\end{tabular}
\end{center}
\caption{Effect on the different loss functions.
The SSIM-based loss function has faster convergence speed and better enhancement performance.
}
\label{fig: Effect on Different Loss Functions}
\end{figure}

\begin{table}[!t]
\centering
\caption{Effect on the updated radio $r$ between the training generator and discriminators.
}
	\scalebox{0.99}{
		\begin{tabular}{lccccccc}
\toprule

 ~~~~~$r$ ~~~~~ & ~~~~~1 ~~~~~  &  ~~~~~2 ~~~~~  &   ~~~~~3 ~~~~~& ~~~~~5  ~~~~~&   ~~~~~10 ~~~~~
\\
\toprule
PSNR $\uparrow$  &19.84           &19.67 &19.62  & 19.80 & 19.43                \\
SSIM $\uparrow$  & 0.8186      &0.8185 &0.8162     & \textbf{0.8234} & 0.8164         \\
\bottomrule
\end{tabular}}
\label{tab:Effect on the updated radio}
\end{table}

\begin{table}[!t]
\centering
\caption{Effect on the hyper-parameter $\alpha$.
}
	\scalebox{0.99}{
		
		\begin{tabular}{lccccccc}
\toprule
$\alpha$ & 0.001  & 0.01 &  0.1 & 1 &  5&  10  \\
\toprule
PSNR $\uparrow$  &19.43           &19.62   & \textbf{19.80} & 19.44&   18.87&19.69                \\
SSIM $\uparrow$  & 0.8181      &0.8189      & \textbf{0.8234} & 0.8176&0.8111&0.8165         \\
\bottomrule
\end{tabular}}
\label{tab:Effect on the Hyper-parameter alpha.}
\end{table}
\begin{table}[!t]
	\centering
\caption{Effect on the hyper-parameter $\beta$.
}
	\scalebox{0.99}{
	
		\begin{tabular}{lccccccc}
\toprule
$\beta$ & 0.00001 & 0.0001  &  0.001 & 0.01 &  0.1& 1\\
\toprule
PSNR $\uparrow$  &19.15           &19.23   & 19.80 & 19.88&   20.06&19.81                \\
SSIM $\uparrow$  & 0.8151      &0.8210      & \textbf{0.8234} & 0.8214&0.8184&0.8156         \\
\bottomrule
\end{tabular}}
\label{tab:Effect on the Hyper-parameter beta.}
\end{table}
\subsubsection{Effect on the Different Loss Functions}\label{sec: Effect on Different Loss Functions}
As we use SSIM-based loss as our pixel reconstruction loss, we are necessary to analyze its effect compared with traditional $L_{1}$-based and $L_{2}$-based losses.
The comparison results on training curves are presented in Fig.~\ref{fig: Effect on Different Loss Functions}.
We note that SSIM-based pixel reconstruction loss has a faster convergence speed and better enhancement performance.
Hence, we use SSIM-based loss as our pixel reconstruction loss in this paper.

\begin{figure}[!t]
\begin{center}
\begin{tabular}{ccccc}
\includegraphics[width = 1\linewidth]{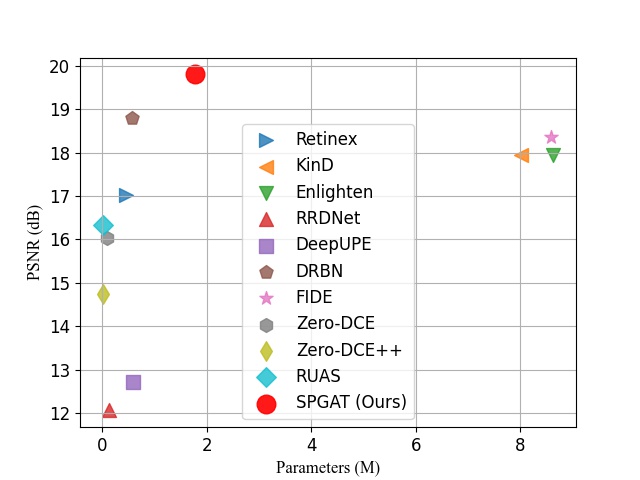}\\
(a) Results on the LOL dataset
\\
\includegraphics[width = 1\linewidth]{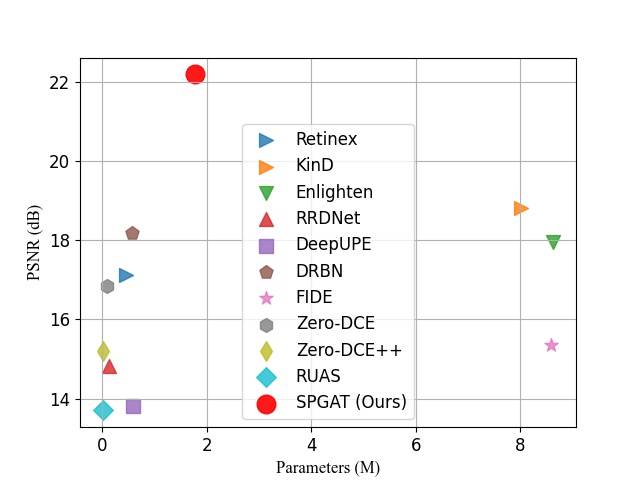}\\
(b)  Results on the Brightening dataset
\\
\end{tabular}
\end{center}
\caption{Parameter-performance trade-off on the LOL and Brightening datasets.
Our method achieves a better trade-off on the two datasets.
}
\label{fig: Parameter-performance trade-off on the LOL and Brightening datasets.}
\end{figure}

\subsubsection{Effect on the Updated Radio $r$ between the Training Generator and Discriminators}\label{sec: Effect on the updated radio $r$ between the training generator and discriminators}

Tab.~\ref{tab:Effect on the updated radio} shows the effect of the updated radio $r$ between the training generator and discriminators.
The model obtains better performance when $r$ is 5.
Hence, $r = 5$ is set as the default setting.

\subsubsection{Effect on the Hyper-Parameters $\alpha$ and $\beta$}\label{sec: Parameter-Performance Comparisons}
Tab.~\ref{tab:Effect on the Hyper-parameter alpha.} and Tab.~\ref{tab:Effect on the Hyper-parameter beta.} respectively report the effect on the hyper-parameter $\alpha$ and $\beta$ in \eqref{eq:total}.
Note that the performance reaches the best when $\alpha$ and $\beta$ are respectively 0.1 and 0.001.
Hence, we set $\alpha = 0.1$ and $\beta = 0.001$ as the default settings.

\subsection{Parameter-Performance Comparisons}\label{sec: Parameter-Performance Comparisons}

In Fig.~\ref{fig: Parameter-performance trade-off on the LOL and Brightening datasets.}, we provide the parameter-performance trade-off comparisons on the LOL and Brightening datasets.
We note that the proposed method achieves a better trade-off in terms of accuracy and model sizes.
\begin{figure}[!t]
\begin{center}
\begin{tabular}{ccccc}
\includegraphics[width = 0.48\linewidth]{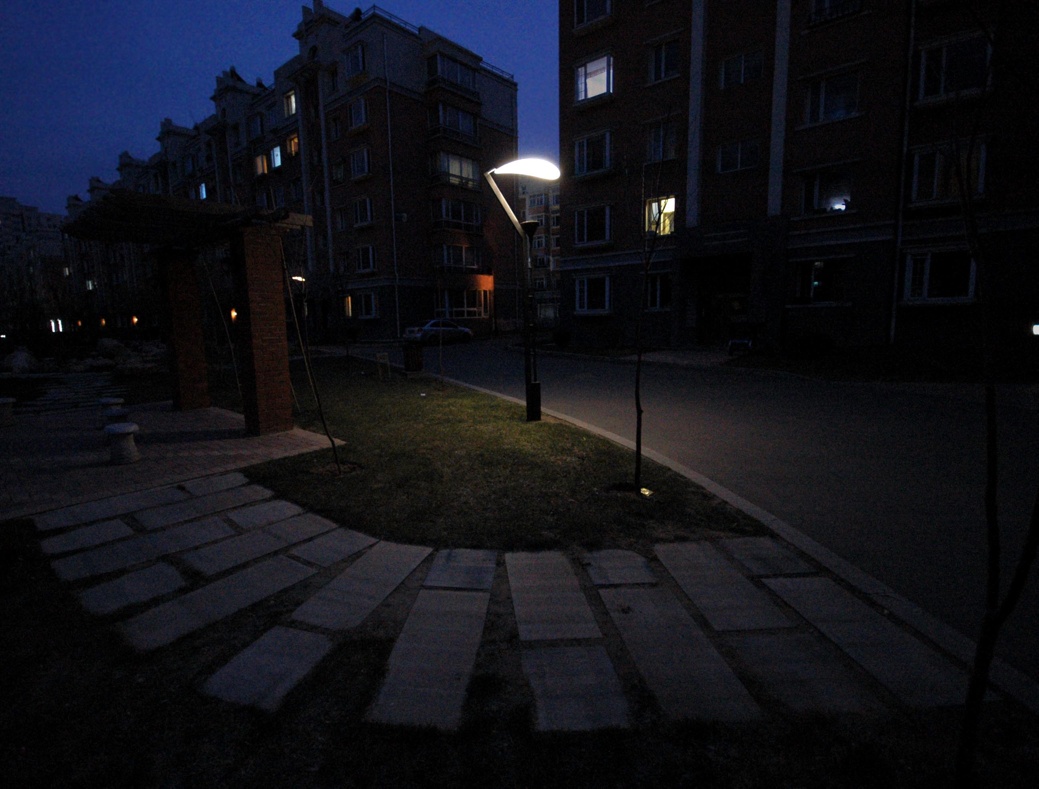}&\hspace{-4.5mm}
\includegraphics[width = 0.48\linewidth]{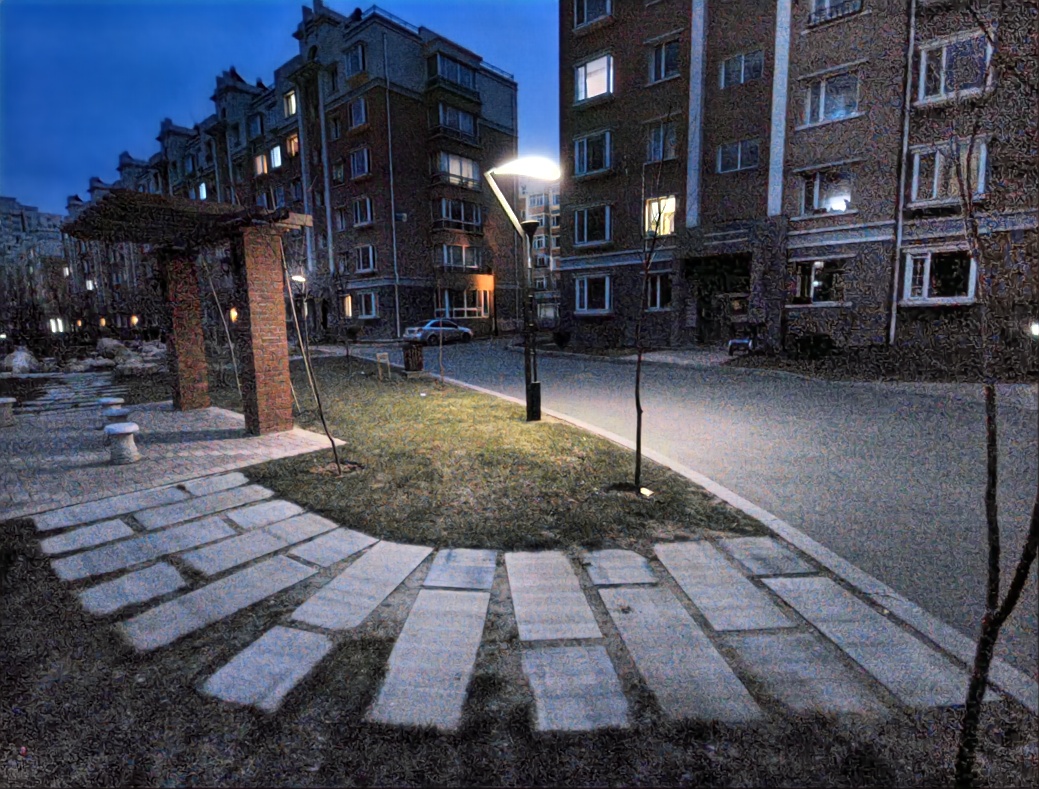}
\\
(a) Low-light &\hspace{-4.5mm}  (b) DRBN~\cite{DRBN_yang_cvpr20}
\\
\includegraphics[width = 0.48\linewidth]{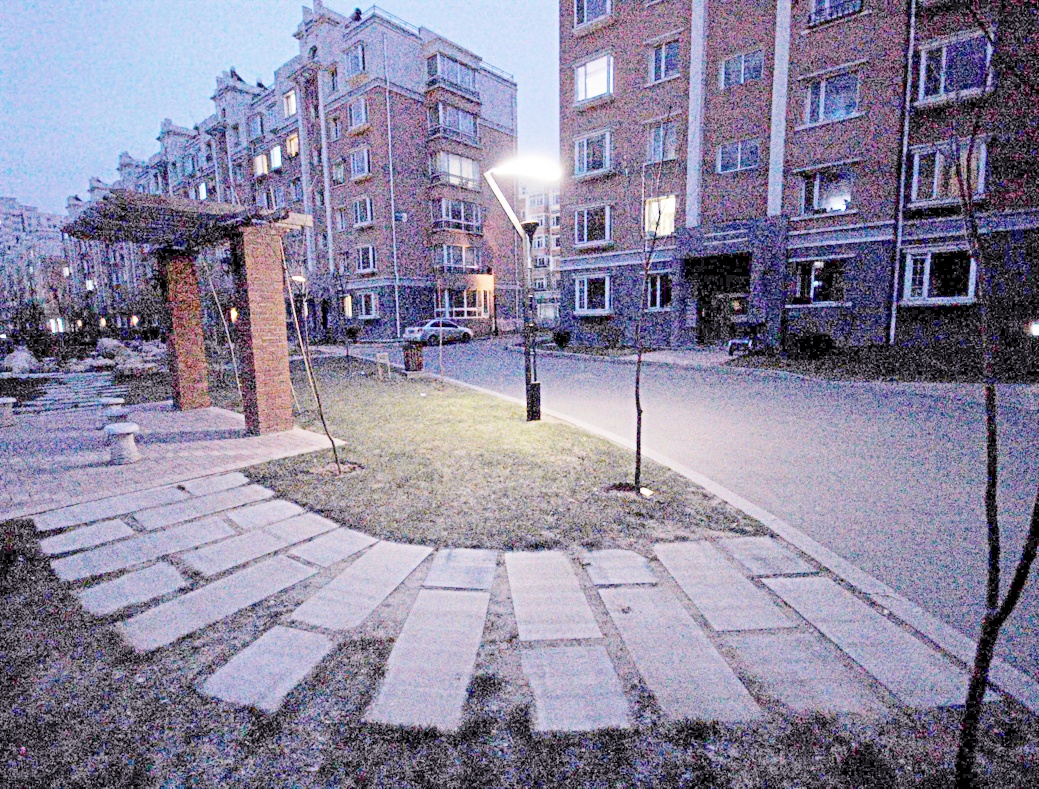}&\hspace{-4.5mm}
\includegraphics[width = 0.48\linewidth]{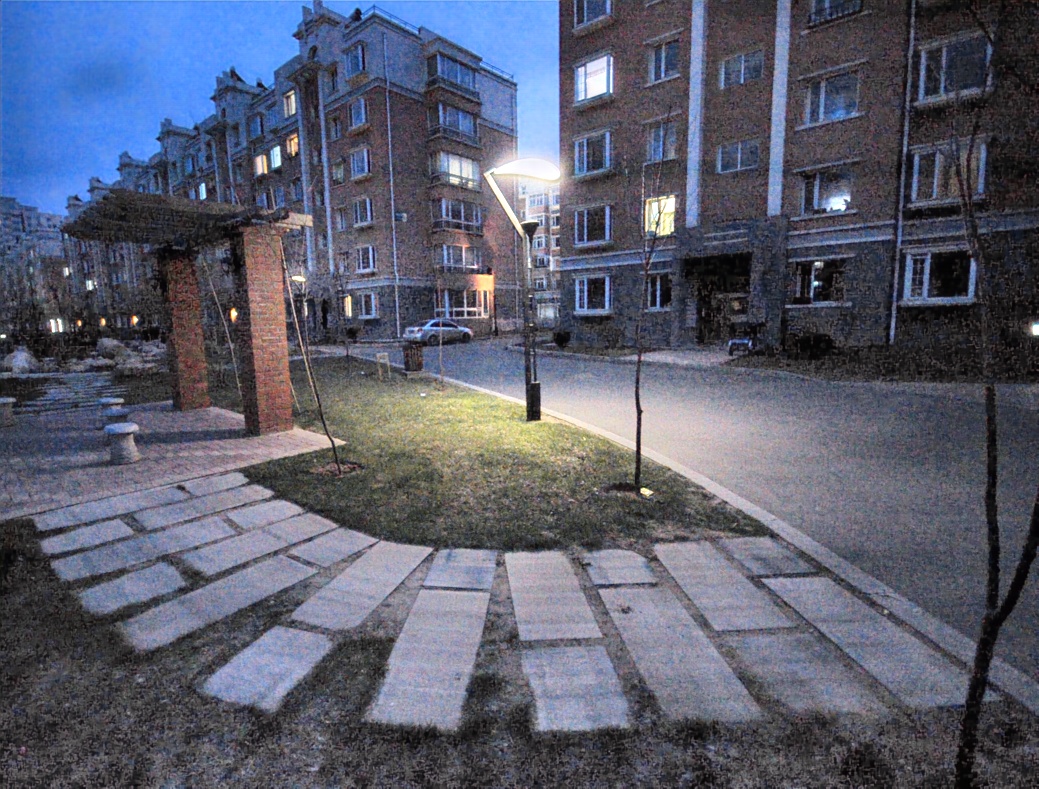}
\\
(c) Zero-DCE~\cite{zerodce_lowlight_guo} &\hspace{-4.5mm}  (d) SPGAT
\\
\end{tabular}
\end{center}
\caption{Limitations.
Our SPGAT as well as state-of-the-art methods hand down some noises when handling extreme low-light degradations.
}
\label{fig: Limitations.}
\end{figure}
\subsection{Limitations}\label{sec: Limitations}

Although our method can generate more natural enhancement results with finer structures, it has some limitations.
Fig.~\ref{fig: Limitations.} shows that our method can not handle the case of extreme low-light degradations well.
Our approach as well as state-of-the-art methods hand down some noises when handling extreme low-light degradations.
This may be caused by that the synthesized low-light images can not model the real-world low-light conditions well.
We leave this for future research.

\section{Conclusion}\label{sec: Conclusion}
In this paper, we have proposed a Structural Prior guided Generative Adversarial Transformer (SPGAT) for low-light image enhancement.
Our SPGAT is a Transformer-based GAN model, which contains one Transformer generator, two Transformer discriminators, and one Transformer structural prior estimator.
The proposed Transformer generator is built on a U-shaped architecture with skip connections and guided by the structural prior estimator for better enhancement.
Meanwhile, we also have proposed a new discriminative training manner by building the skip connections between the generator and discriminators with the guidance of structural prior.
By designing such a model, our SPGAT is able to produce more natural results with better details.
Extensive experiments have demonstrated that SPGAT achieves superior performance on both synthetic and real-world datasets.
%
%
\bibliographystyle{IEEEtran}
\bibliography{egbib}

\end{document}